\documentclass{article}

\usepackage[english]{babel}
\usepackage{natbib}
\usepackage{subcaption}

\usepackage[letterpaper,top=2cm,bottom=2cm,left=3cm,right=3cm,marginparwidth=1.75cm]{geometry}

\usepackage{amsmath}
\usepackage{rotating}
\usepackage{graphicx}
\usepackage{fullpage}
\usepackage{bm}
\usepackage[colorlinks=true, allcolors=blue]{hyperref}
\usepackage{float}
 
\title{Historical and psycholinguistic perspectives on morphological productivity: A sketch of an integrative approach}
\author{Harald Baayen$^1$, Kristian Berg$^2$, and Mirrah Maziyah Mohamed$^1$ \\ \ \\
University of T\"{u}bingen (1) and University of Oldenburg (2)}

\begin{document}
\maketitle

\section{Introduction}

In this study, we approach morphological productivity from two perspectives: a cognitive-computational perspective,  and a diachronic perspective zooming in on  an actual speaker, Thomas Mann.  For developing the first perspective, we make use of a cognitive computational model of the mental lexicon, the discriminative lexicon model (DLM). For developing the second perspective, we study how the intake and output of one prolific writer changes over time.
    
The DLM models implements mappings between numeric representations for words' forms, and numeric representations for their meanings, using embeddings from distributional semantics. Model mappings can be evaluated not only on how precise their predictions are for the data on which they have been trained, but also how well they predict novel, unseen data.  We will illustrate the promise of using model accuracy on held-out data as a novel fine-grained measure of productivity.  To do so, we present case studies of nominal inflection in Finnish, derivational prefixation in Malay, and compounding in English. 
    
Although the DLM is a model for a language user's mental lexicon, a DLM model is typically trained using words as observed in corpora.  Typically, these corpora sample language from many different speakers. Because individual speakers have their own areas of interest and expertise, they will know the vocabulary specific to these interests, but will be less familiar or unfamiliar with the vocabulary of other areas of specialization.  As a consequence, corpora overestimate the vocabularies of individual speakers, and the ecological validity of computational models targeting an individual's mental lexicon are inevitably negatively affected when trained on aggregated community data. 

We therefore complement this cognitive-computational perspective with a socio-historical perspective, investigating the productivity of selected German derivational suffixes in the writings of Thomas Mann,  comparing what he is known to have read (community input) with his own writings (personal output).  We will sketch how a DLM model could be set up for Thomas Mann, and present some first findings.

A common thread across all our case studies is the use of embeddings to represent words' meanings. We will show that for inflection and derivation, the DLM model links n-grams that are co-extensive with inflectional or derivational exponents with the most prototypical embeddings of the words that use these exponents.  We will also show that Thomas Mann is more likely to produce a novel German derived word with a given suffix when the embeddings of the words with that suffix cluster more closely around the prototypical embedding.  

In what follows, we first report our computational experiments with the DLM as a tool for gauging productivity. We then turn to our study of Thomas Mann and the productivity of German suffixes in his writings.  We conclude with a discussion of our findings. 

\section{Productivity: approximations with the DLM}

\subsection{General considerations}

The discriminative lexicon model (DLM) is a computational tool for, on the one hand, assessing how well mappings between form and meaning can be learned, and on the other hand, for tracing the consequences of learnability for lexical processing.  The model requires the analyst to define mappings between form and meaning for comprehension. Both words' forms and words' meanings are `embedded' in high-dimensional vector spaces.  The numeric vectors for words' forms are brought together as the row vectors of a matrix (table) $\bm{C}$, and the numeric vectors for words' meanings are brought together as the row vectors of a matrix $\bm{S}$.  For comprehension a mapping $f$ takes the form vectors and transforms them into the corresponding meaning vectors, as precisely as possible:
$$
f(\bm{C}) = \bm{S}.
$$
For comprehension, a mapping $g$ takes the semantic embeddings in $\bm{S}$ and transforms these into their form embeddings $\bm{C}$:  
$$
g(\bm{S}) = \bm{C}.
$$ 
As is customary in machine learning, model parameters (which define the mappings $f$ and $g$) are estimated for a training dataset, and the quality of the model is evaluated on held-out test data. This provides a natural framework for evaluating morphological productivity, namely, by asking the question how well the model is able to understand and produce novel complex words in the held-out dataset, i.e., words that it has not encountered before during training. 

For mappings between form and meaning to be productive, in the sense that novel, previously unencountered words, can be understood and produced, there must be systematicities between the form space and the semantic space. If the relation between form and meaning would be truly arbitrary, a model could memorize form and meaning pairings, but there is no way in which the model would be able to generalize to novel test data.  Thus, the arbitrariness of the linguistic sign \citep{DeSaussure:66} is a worst-case scenario for any learning algorithm.  Morphology, however, breaks this arbitrariness.  What is of interest to us here is that the way in which morphology breaks this arbitrariness differs substantially between inflection, compounding, and derivation, and that the consequences for generalization are profound.   

In the literature on systemic productivity, restrictions (on the form and meaning of base words, and the form and meaning of complex words) have played a foundational role \citep[see, e.g.,][]{Booij:77}.  We maintain that such restrictions enable generalization. Without such restrictions, generalization is impossible. We will argue, and present modeling evidence, that compounding, lacking such restrictions, does not enable solid generalization, whereas inflection and derivation, thanks to being constrained by restrictions, do enable generalization.

In what follows, we present a series of computational experiments that clarify this perspective on morphological productivity.  We first zoom in on inflection, using nominal inflection in Finnish as a non-trivial example of the issues that any theory of morphology seeking to predict inflectional productivity has to solve.   We then discuss the very opposite kind of word formation, compounding, using datasets from English, but briefly venturing into compounding in Mandarin Chinese.  Following this, we target derivational morphology, using prefixal morphology in Malay as our working example.

\newpage

\subsection{Inflection}

Inflectional morphology is where regularity tends to be most clearly visible.  Yet, inflectional systems often have their own pockets of irregularity. Past-tense inflection in English provides a well-known example. Regular past-tense inflection with the dental suffix has been described (and hotly debated) as fully regular and productive, and the irregular past tense forms as being unproductive and stored in memory.  In this section, we consider a very different inflectional system: nominal inflection in Finnish.  Finnish nouns are inflected for two numbers and fourteen cases.  At the form side,  some 49 inflectional classes are distinguished \citep[see][for detailed discussion]{Nikolaev:Chuang:Baayen:2025}.  At the semantic side, \citet{nikolaev:2022} have shown that the change in semantic space from singular to plural varies systematically by case. In other words, number and case do not have straightforward simple semantic main effects, instead, number and case enter into a semantic interaction.   

Can the DLM master this complex inflectional system, without having recourse to stems, features for stem allomorphy, features for inflectional class, and rules of referral (most plural forms are based on the partitive singular)?  \citet{Nikolaev:Chuang:Baayen:2025} tested the DLM on 55,271 nominal forms associated with the 2000 most frequent Finnish nouns.  Using fasttext embeddings to represent words' meanings, 4-gram vectors to represent words' forms,  and frequency-informed learning, they observed that on the training data, the model correctly understood 75.96\% of the types and 95.76\% of the 40,694 tokens of the dataset on which the model was trained. For held-out data, the 14,577 words with a frequency less than or equal to five, 64.29 were correctly understood. For 98.29\% of the held-out tokens, the predicted vector was among the top 10 closest neighbors of the targeted `gold standard' embedding.  These results indicate that the DLM comprehension model is productive, in that it can understand many inflected forms that it has not encountered during training.  

\begin{figure}[hb]
    \centering
    \includegraphics[width=1.0\linewidth]{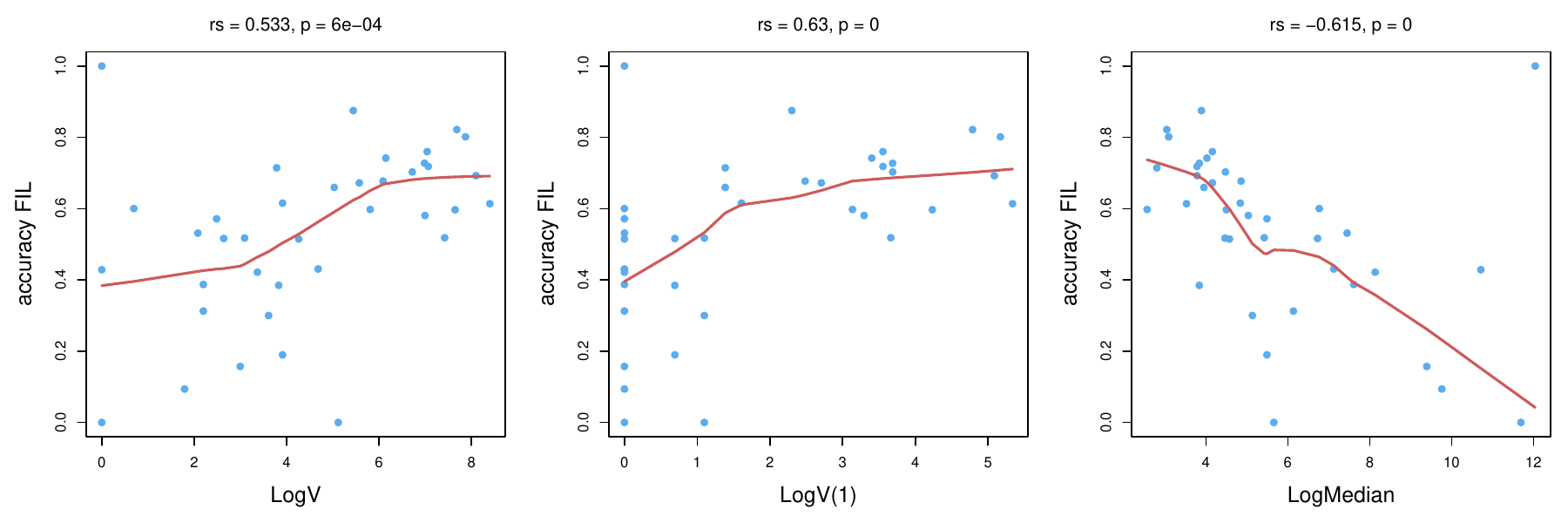}
    \caption{By-class accuracy of a frequency-informed comprehension model on held-out data as a function of three productivity measures. Dots represent inflectional classes.}
    \label{fig:finnish_scatterplots}
\end{figure}

However, this overall assessment of the DLM's performance (and by implication, the productivity of the Finnish noun system) is not fair, as the 49 inflectional classes differ substantially in size and productivity.  Some inflectional classes have many types, many hapax-legomena, and low median word frequency.  Others comprise only small numbers of types, few if any hapax legomena, and many high-frequency words.  Performance of the DLM on the less productive and unproductive inflectional classes should be worse than its performance on more productive inflectional classes.  This is exactly what \citet{Nikolaev:Chuang:Baayen:2025} observed, as illustrated in Figure~\ref{fig:finnish_scatterplots}. As the number of types in an inflectional class increases, the accuracy of the DLM on held-out data increases as well (left panel). Similarly, inflectional classes with more hapax legomena afford greater DLM accuracy, again on held-out data (center panel). Furthermore, as the median word frequency in an inflectional class increases, prediction accuracy for held-out data decreases.  

We propose that the prediction accuracies of the DLM for the different inflectional classes provide insight into the degrees of productivity of these classes. These usage based, learning-driven accuracies integrate the joint effects of type and token frequencies, as well as the effects of idiosyncracies at the level of word form and the level of word meaning.

\begin{sidewaysfigure}
\centering
    \includegraphics[width=0.3\linewidth]{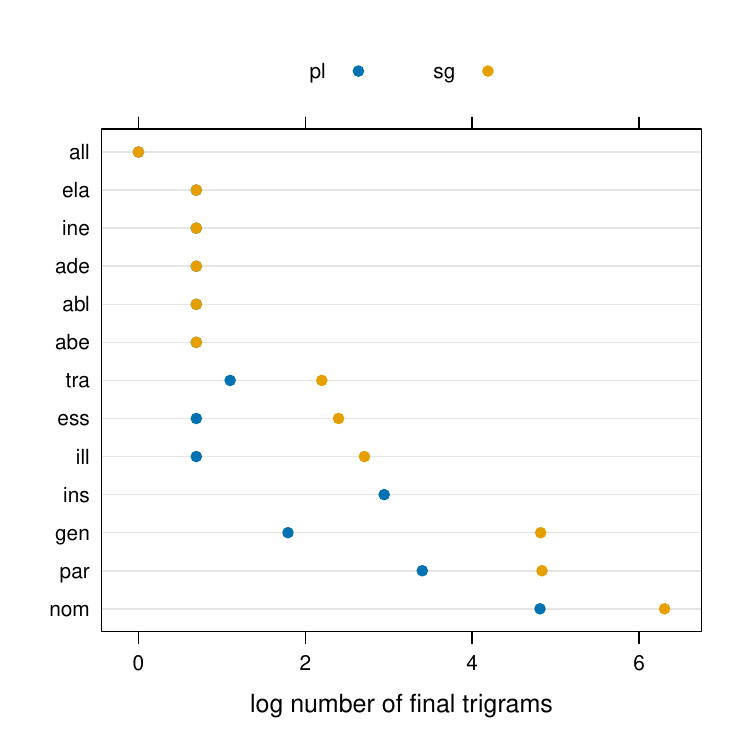}\includegraphics[width=0.3\linewidth]{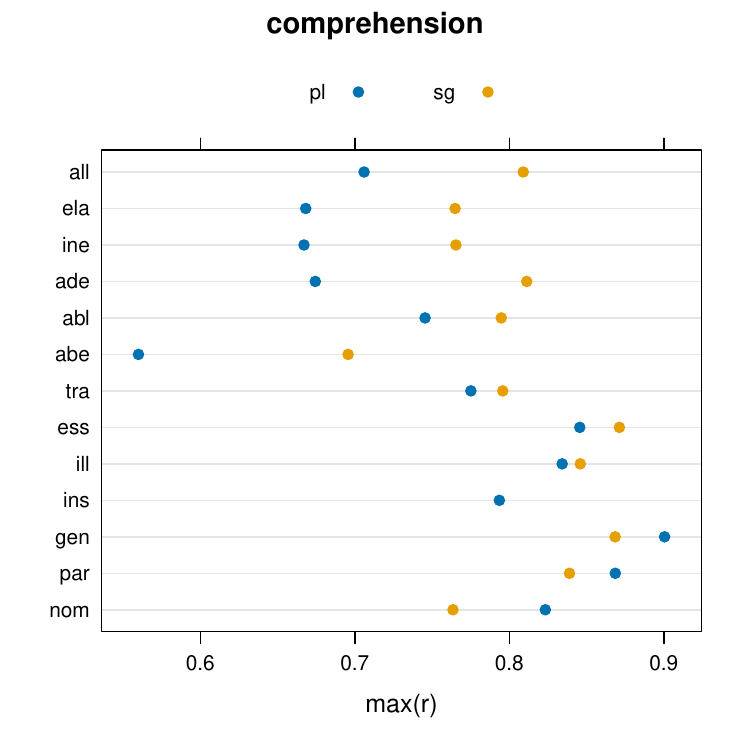}\includegraphics[width=0.3\linewidth]{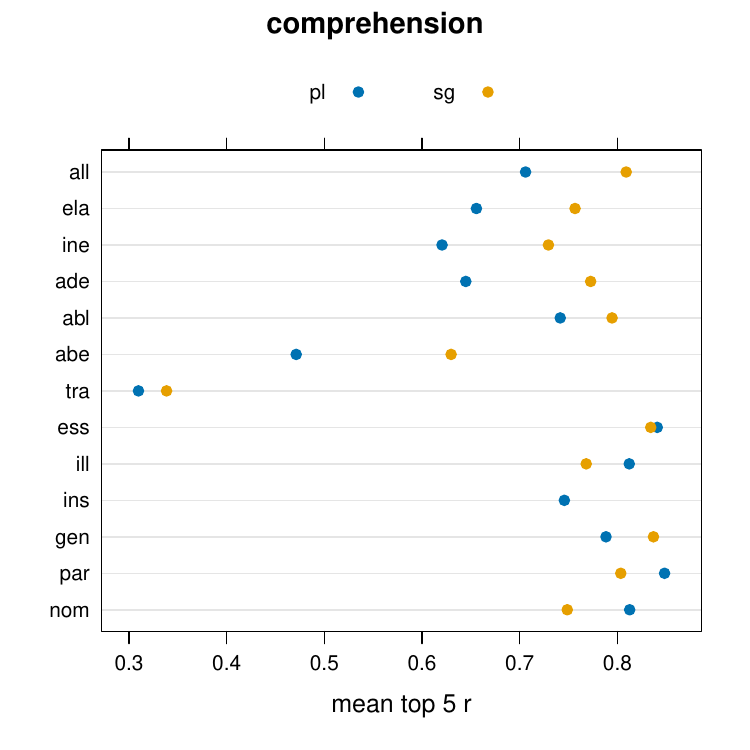}
    \includegraphics[width=0.3\linewidth]{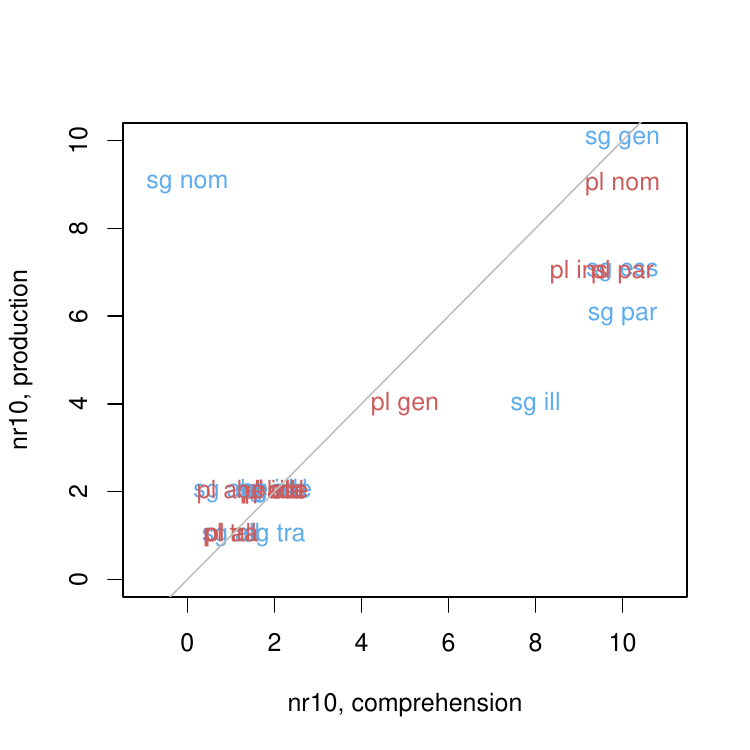}\includegraphics[width=0.3\linewidth]{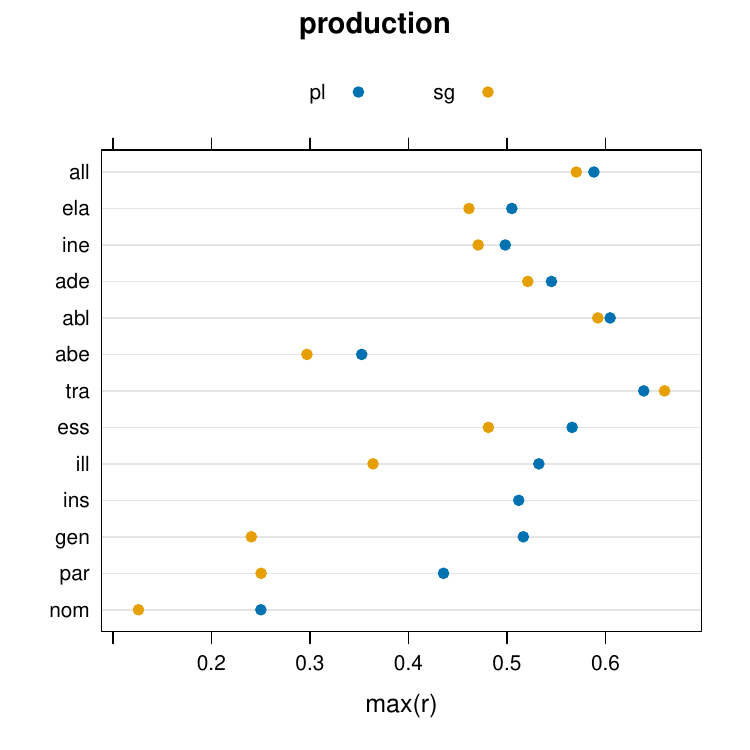}\includegraphics[width=0.3\linewidth]{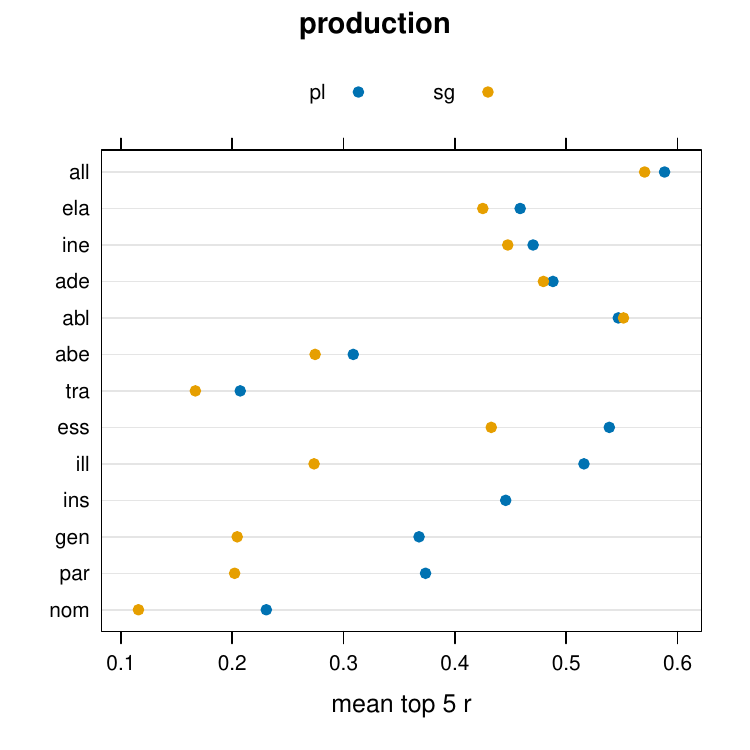}   
    \caption{Row vectors of $\bm{F}$ and column vectors of $\bm{G}$ in relation to the centroids of case-number combinations in Finnish. Number of final trigrams: word-final three-letter sequences that realize a given case-number combination; max(r): the maximum correlation with the centroid; mean top 5 r: mean of the top-five ranked correlations; nr10: the number of final trigrams among the top 10 highest correlations.}
    \label{fig:finnish_centroids_comprehension}
\end{sidewaysfigure}

Thus far, we have used the DLM as a black box. We have shown that the black box exhibits the theoretically expected behavior. But why does it work so well?  In what follows, we will pry open the black box, and show that what the model does is remarkably well-interpretable, and is surprisingly similar to realizational theories of morphology. 

Recall that the DLM sets up a matrix $\bm{C}$ of form vectors. Each row of such a table specifies which n-grams are present in a word's form.  For the Finnish words \textit{vuonna} (`year', a lexicalized form of the singular essive) and \textit{kello} (`clock'), the `form embeddings' based on letter 4-grams could look like this:
\begin{center}
\begin{tabular}{|l|rrrrrrrrrr|} \hline
       & \#vuo & vuon & uonn & onna & nna\# & \#kel & kell & ello & llo\# & \ldots \\ \hline 
vuonna &  1    & 1    & 1    & 1    & 1     &  0    & 0    & 0    & 0  & \ldots   \\
kello  &  0    & 0    &  0   & 0    & 0     &  1    & 1    & 1    & 1  & \ldots   \\ \hline
\end{tabular}
\end{center}
Only a small part of these embeddings are shown. There are 13,364 different 4-grams in the dataset of Finnish analyzed by \citet{Nikolaev:Chuang:Baayen:2025}, for \textit{vuonna}, $13,364-5=13,359$ 4-grams, the value in the row vector is zero.  The DLM assumes that there is a function $f$ that takes these form embeddings and transforms them into semantic embeddings.  In other words, $f$ applied to the row vectors of the form embeddings $\bm{C}$ returns the corresponding set of semantic embeddings, which we bring together as the row vectors of a table (matrix) $\bm{S}$:
$$
f(\bm{C}) = \bm{S}.
$$
In what follows, we assume that this function $f$ is very simple and implements a linear transformation:
$$
\bm{C}\bm{F} = \bm{S}.
$$
How linear transformations work is explained in detail in \citet{Heitmeier:Chuang:Baayen:2024}. Important for the present discussion is that the row vectors of the table (matrix) $\bm{F}$ specify 4-gram specific embeddings.  The first five rows of the following display show the first 6 cells of these 300-element long 4-gram embeddings.   
\begin{center}
\begin{tabular}{|l|rrrrrrr|} \hline
\#vuo & -0.007273 &-0.014577 &-0.026716 &0.014613 &-0.004810 & -0.019703 & \ldots \\
vuon & -0.032234 &-0.023539 & 0.003362 &0.001231 & 0.008429  &0.022490 & \ldots \\
uonn &  0.030358 &-0.015991 &-0.010521 &0.022888 &-0.001061  &0.032784 & \ldots \\
onna & -0.000150 & 0.007794 & 0.011207 &0.003758 &-0.009883 & -0.002486 & \ldots \\ 
nna\# & 0.007692 & 0.010210 & -0.007754 & -0.006202 & -0.037584 & -0.018848 & \ldots \\ \hline
\textsc{sum} &   -0.001606 & -0.036103 & -0.030423 &   0.036290 & -0.044910 &  0.014237 & \ldots \\ \hline
\end{tabular}
\end{center}
What the linear mapping $\bm{F}$ does is look up in table $\bm{C}$ which 4-grams occur in say \textit{vuonna}, retrieve the corresponding 4-gram specific embeddings, and sum these column wise. The first 6 elements of the resulting semantic embedding predicted for \textit{vuonna} are shown in the last row of the above display.   The mapping $\bm{F}$ faces a hard task: it has to stay faithful to the meaning of the lexeme, and not confuse the meaning of \textit{vuosi} (`year') with the meaning of \textit{kello} (`clock').  It has to realize essive singular for \textit{vuonna}, and nominative singular for \textit{kello}.  And it has to make peace with the extensive stem allomorphy that characterizes Finnish nouns (for \textit{vuosi}, \textit{vuo-, vuon-, vuot-}). The 4-grams that are part of the stem have to take care of implementing the meaning ``year''.  The initial 4-gram \textit{\#vuo} is an excellent cue for ``year'', all forms of \textit{vuosi} start with \textit{\#vuo}, and there are only two other lexemes in the dataset that share this initial trigram: \textit{vuokra} `rent' and \textit{vuokralainen}  `tenant'. The next 4-gram, \textit{vuon}, is unique to \textit{vuosi}, but occurs in only four forms. The remaining 4-grams incorporate part of the inflectional endings, which are shared with many other words.  What semantic embeddings does the linear mapping estimate for these trigrams? To address this question, we inspected the final 4-grams, which consist of three letters and the word boundary marker \textit{\#}.  We limited our exploration to the 34,262 words in our dataset that have no possessive clitics nor discourse clitics.

Given that in the semantic system of Finnish, case and number interact \citep{nikolaev:2022}, we collected all word-final 4-grams (e.g, \textit{nna\#} for \textit{vuonna}) and sorted them by the case+number combinations.  The number of word-final 4-grams for the case+number combinations ranges from 1 (for allative singular and allative plural) to 547 (for the nominative singular).  The upper left panel of Figure~\ref{fig:finnish_centroids_comprehension} presents these counts (log-transformed) for the different cases, broken down by number.  The nominative clearly pays a high price for being unmarked. The other two grammatical cases, partitive and genitive, also have large numbers of final endings, reflecting that these cases often have somewhat unpredictable stems. 

\begin{figure}[htbp]
    \centering
    \includegraphics[width=0.6\linewidth]{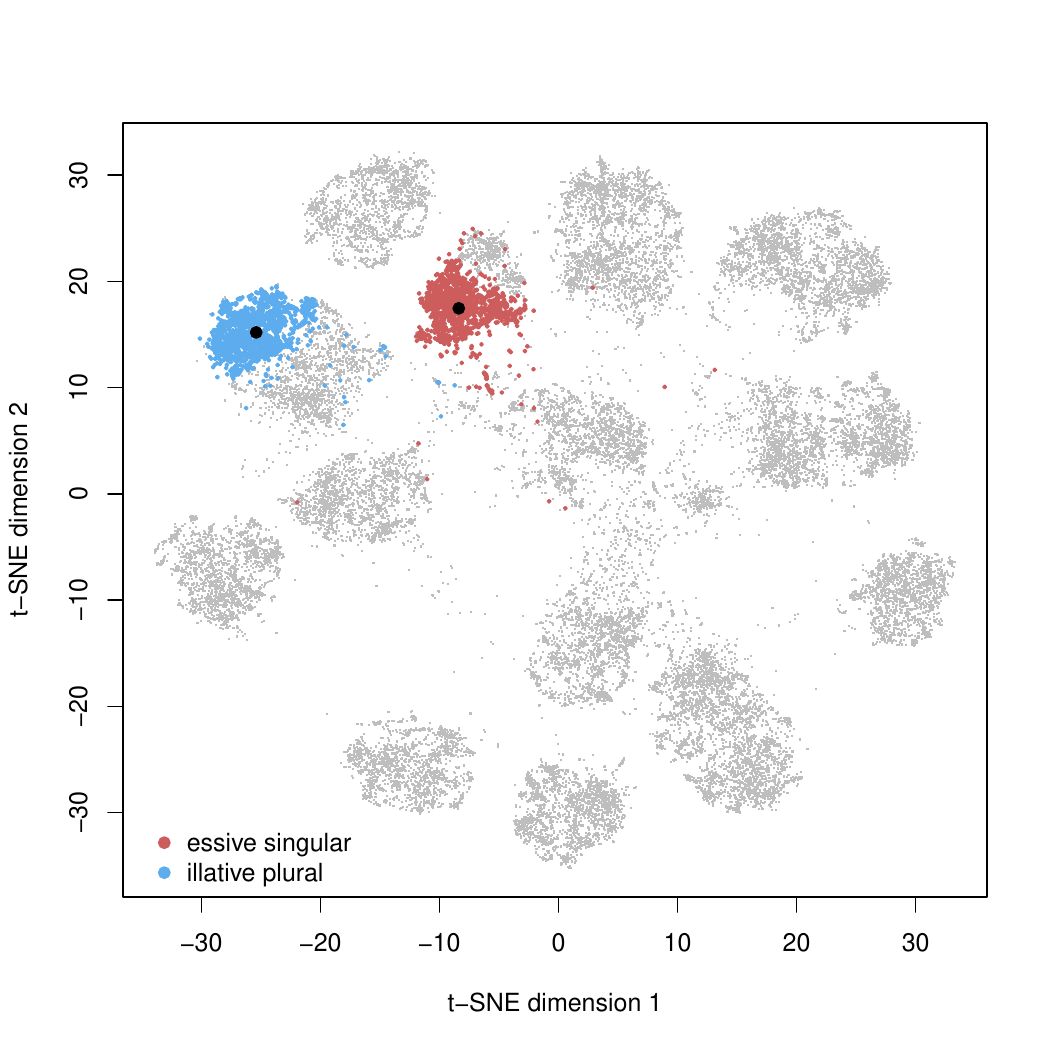}
    \caption{Position of 34,262 Finnish nominal forms in a 2-dimensional t-SNE plane. Clustering is by case. The cluster for essive singular is presented in red, and the cluster for illative plural is given in blue. The black points represent the centroids of these clusters.}
    \label{fig:tsneFinnish}
\end{figure}

For each of the word-final 4-grams, there is a row in the $\bm{F}$ mapping that represents the semantic contribution of that n-gram to its carrier word. Our hypothesis is that this semantic contribution must be similar to the centroid of the case-number combination that the 4-gram is tied to. Figure~\ref{fig:tsneFinnish} illustrates the centroids for essive singulars (in red) and illative plurals (in blue), using the t-SNE unsupervised clustering algorithm \citep{maaten2008visualizing} to re-represent the embeddings of the inflected nouns in a 2-dimensional space. The larger clusters represent case, within clusters, plurals and singulars form sub-clusters. The centroids are in the center of the cluster, and represent the most prototypical meaning of a case-number combination.  Our intuition is that the best the linear mapping can do is use embeddings for final 4-grams that are similar to the centroid of the case-number cluster that these 4-grams have to realize.  

\newpage

\begin{table}[ht]
\centering
\caption{
Correlations of case-number centroid and final 4-gram embedding (row vector of $\bm{F}$) and rank in the ordered sequence of all 4-grams with this centroid.  
}
\label{tab:finnishRanks}
\begin{tabular}{lrr} \hline
\multicolumn{3}{c}{illative plural} \\ \hline
 4-gram \hspace*{3em} & \hspace*{3em}        r &  \hspace*{3em} rank \\ \hline
 iin\#  & 0.83  &    1   \\
 hin\#  & 0.79 &    2   \\ \hline
\multicolumn{3}{c}{essive singular} \\ \hline
 4-gram &         r &   rank \\ \hline
una\#   & 0.87 &     1   \\
ana\#   & 0.86 &     2   \\
ena\#   & 0.85 &     3   \\
enä\#   & 0.81 &     4   \\
ona\#   & 0.77 &     5   \\
ina\#   & 0.76 &     6   \\
inä\#   & 0.73 &     7   \\
önä\#   & 0.72 &     8   \\
änä\#   & 0.71 &     9   \\
ynä\#   & 0.62 &    10   \\
nna\#   & 0.18 &  1966   \\ \hline
\multicolumn{3}{c}{nominative singular} \\ \hline
 4-gram &         r &   rank \\ \hline
kka\#   & 0.76 &  11 \\
eri\#   & 0.76 &  12 \\
lma\#   & 0.75 &  19 \\
eli\#   & 0.75 &  20 \\
uri\#   & 0.72 &  38 \\
ari\#   & 0.72 &  47 \\
\vdots  & \vdots    & \vdots \\ \hline
\end{tabular}
\end{table}

The reason is straightforward: the centroid averages out the differences in the consequences of the very different affordances of the referents of individual nouns and their consequences for the corresponding embeddings.  Because a given word-final 4-gram has to ``serve'' many different nouns simultaneously, an embedding that is similar to the centroid, the average embedding, is the best solution.

To put this line of reasoning to the test, we calculated the correlation matrix of the centroids with the row vectors of the mapping $\bm{F}$.
The following display shows a small part of this 26 $\times$ 13,364 matrix.
\begin{center}
\begin{tabular}{l|rrrrrrr} \hline
 & \#vuo & vuon & uonn & onna & nna\# & \#kel & \ldots \\ \hline
sg ess & 0.45 & 0.12 & 0.19 & -0.20 & 0.18 & 0.33 & \ldots \\ 
sg nom & 0.55 & -0.12 & 0.11 & -0.00 & 0.30 & 0.58 & \ldots \\ 
sg ill & 0.26 & -0.00 & 0.08 & 0.01 & 0.15 & 0.39 & \ldots\\ 
sg gen & 0.42 & 0.00 & 0.14 & -0.08 & 0.15 & 0.42 & \ldots\\ 
sg tra & 0.35 & 0.06 & 0.20 & -0.12 & 0.12 & 0.35 & \ldots\\ 
sg par & 0.46 & -0.04 & 0.07 & -0.06 & 0.13 & 0.46 & \ldots\\ 
\vdots & \vdots &\vdots &\vdots &\vdots &\vdots &\vdots & $\ddots$ \\ \hline
\end{tabular}
\end{center}
If our intuition is correct, we expect that the 4-grams with the highest correlations with a given centroid will be the word-final 4-grams that occur in words that realize the relevant combination of number and case.  Table~\ref{tab:finnishRanks} presents results for three case-number combinations: illative plural, essive singular, and nominative singular. For the illative plural, there are only two word-final 4-grams, and their embeddings are well-correlated with the centroid of the illative plural cluster. The correlations, 0.83 and 0.79, are the most correlated of all 13,364 4-grams (ranks 1 and 2). The essive singular is realized by a wider range of 4-grams. For 10 4-grams, the correlations are high and occupy ranks 1-10. It is only the 4-gram \textit{nna\#} (as in \textit{vuonna}) that is not well correlated with its centroid: many irrelevant 4-grams (1955) have higher correlations.  For the unmarked nominative singular, which is associated with 547 final 4-grams, correlations are lower, and ranks are higher.   

The upper central panel of Figure~\ref{fig:finnish_centroids_comprehension} presents the maximal correlation of a word-final 4-gram with its case-number centroid. Except for the nominative singular and the abessive plural, this maximal correlation is always associated with an appropriate word-final 4-gram.  In the same figure, the upper right panel presents the mean of the correlations of the 5 most highly ranked correlations. The two panels provide solid support for the idea that the semantics contributed by the word-final 4-grams, the fuzzy `exponents' of the DLM model, realize semantics that are highly similar to the centroid of their corresponding case-number embeddings.

For the mapping required for production, $\bm{G}$, which starts out with the semantic vectors and uses these to predict words' 4-grams, a similar pattern is observed. It is now the the column vectors of $\bm{G}$, which are used to support 4-grams, that are strongly correlated with the centroids of the case-number clusters. As can be seen by comparing the center and right lower panels in Figure~\ref{fig:finnish_centroids_comprehension} with their upper panel counterparts, the correlations are somewhat lower for production as compared to comprehension. There are four plural cases for which the best correlation is not ranked first.  This reduced correlations with the centroids are likely due to the mapping from meaning to form (300 dimensional meaning vectors to 13,364 dimensional form vectors) cannot be very precise as it is not possible to map from a lower-dimensional space into a higher-dimensional space with precision.  The lower left panel of Figure~\ref{fig:finnish_centroids_comprehension} presents the count of word-final 4-grams that have ranks of at most 10.  The nominative singular is the outlier in this plot: it has good ranks for production, but not for comprehension.

In summary, we have argued that the accuracy of the DLM model on held-out data may be useful as a measure of productivity.  This measure lines up well with other already well-established measures of productivity, and elegantly navigates the complexities of the inflectional classes of the Finnish noun system.  Furthermore, we have shown that the mappings of the DLM associate word-final 4-grams with the prototypical embeddings of case-number combinations. These centroids form the basis of generalization. Because the centroids are means, they remain fairly stable even when word forms are withheld from the training data, and hence enable generalization to novel forms on which the model was not trained.  Although the DLM does not attempt to isolate stems and exponents, and works with multiple word-final 4-grams for a given case-number combination,  these 4-grams come very close to `realizing' the prototypical semantics of case and number.

\subsection{Compounding}

How productive is compounding? A perusal of the CELEX database \citep{Baayen:CD95} shows that of all entries in the lemma list for English that are analysed as having two constituents, 7.1\% are prefixed words, 42.7 are suffixed words, and 36.9 are compounds.  
Clearly, compounding is a productive word formation pattern.  There are languages such as Vietnamese and Chinese that heavily rely on compounding, with no inflection and hardly any derivation. And yet, in structuralist analyses, compounds are not morphological categories, i.e., sets of words that share aspects of form and aspects of meaning \citep[see, e.g.,][]{Schultink:1961}.  If compounds have no clear systematic relations between form and meaning to lay the basis for generalization, how then can they be productive?    

To address this question, we carried out a computational experiment with 11,170 English words, split into training data (10959 words, 7301 compounds (written as one word), 3658 constituents) and test data (811 compounds with constituents that occur in the training data). Comprehension accuracies for endstate learning (EOL) and frequency informed learning FIL) are listed in Table~\ref{tab:compound_accuracies} for training data and test data, for accuracy (@1) and loosely evaluated accuracy (being among the top 10, @10).  For the training data, accuracies are high for the training data (and for FIL, as expected, accuracy is only high using token-wise evaluation).  For the held-out test data, all compounds with known constituents, accuracy @1 plummets to 12.6\% for EOL, and to zero for FIL. The higher score for EOL (but not FIL) for lenient evaluation @10 for the novel compounds, 69.1\%, suggest that the model may get the gist of the semantics of the novel compounds, but lacks precision.   

\begin{table}[htbp]
    \centering
    \caption{Comprehension accuracies for training and test data. For the training data, token-wise accuracy @1 for FIL is 0.858.}
    \label{tab:compound_accuracies}
\begin{tabular}{lrrrr} \hline 
           & \multicolumn{2}{c}{training} & \multicolumn{2}{c}{testing} \\ \hline 
  learning &  @1     & @10    &  @1    &   @10 \\ \hline
  EOL      &  0.866  & 0.989  & 0.1258 & 0.6905 \\
  FIL      &  0.079  & 0.691  & 0      & 0.1134 \\ \hline  
\end{tabular}
\end{table}

\begin{table}[ht]
\centering
\caption{The trigrams of \textit{airfield} and the correlation of their row vectors in $\bm{F}$ with the embeddings of selected words. Trigrams spanning the constituent boundary are highlighted in red.
}
\label{tab:airfield}
{\footnotesize
\begin{tabular}{r|rrrrrrrrrr} \hline
 & about & abouts & absorption & absorption-line & abundance & access & \textbf{air} & \textbf{airfield} & craft & \textbf{field} \\ \hline
  air  & -0.099 & -0.067 & -0.070 & -0.040 & -0.016 & -0.034 & \textbf{0.063} & \textbf{0.029} & -0.032 & \textbf{-0.147} \\ 
  ld\# & 0.200 & 0.232 & 0.152 & 0.083 & 0.190 & 0.171 & \textbf{0.318} & \textbf{0.128} & 0.377 & \textbf{0.246} \\ 
  \#ai & 0.130 & 0.126 & 0.217 & 0.200 & 0.091 & 0.139 & \textbf{0.331} & \textbf{0.314} & 0.204 & \textbf{0.232} \\ 
  {\color{red} irf}  & -0.082 & 0.020 & -0.142 & -0.085 & -0.025 & 0.004 & {\color{red}\textbf{-0.145}} & {\color{red}\textbf{0.123}} & -0.120 & {\color{red}\textbf{-0.084}} \\ 
  {\color{red}rfi}  & 0.059 & 0.151 & -0.107 & -0.077 & -0.112 & -0.063 & {\color{red}\textbf{0.030}} & {\color{red}\textbf{0.229}} & -0.021 & {\color{red}\textbf{-0.067}} \\ 
  fie  & -0.137 & 0.005 & -0.141 & 0.047 & 0.035 & -0.095 & \textbf{-0.054} & \textbf{0.103} & -0.084 & \textbf{0.131} \\ 
  iel  & 0.096 & 0.001 & 0.114 & 0.049 & -0.093 & 0.059 & \textbf{-0.069} & \textbf{0.012} & -0.057 & \textbf{0.046} \\ 
  eld  & -0.029 & -0.066 & -0.016 & -0.111 & 0.059 & 0.018 & \textbf{0.087} & \textbf{-0.042} & 0.090 & \textbf{-0.119} \\ \hline
\end{tabular}
}
\end{table}

In order to understand how the model learns to understand the compounds that it has been trained on, it is helpful to consider how the trigrams of a compound contribute to its meaning, using their trigram-specific embeddings in the $\bm{F}$ matrix. In what follows, we set aside the effects of usage, and inspect the $\bm{F}$ mapping obtained with endstate learning.

Table~\ref{tab:airfield} shows a small part of the correlation matrix for the row vectors of $\bm{F}$ and the embeddings in $\bm{S}$. The trigrams shown are those of \textit{airfield}.  The trigrams that straddle the  constituent boundary between \textit{air} and \textit{field}, \textit{irf} and \textit{rfi}, are highlighted in red. In what follows, for ease of exposition, when we mention the `correlation of \textit{irf} and \textit{airfield}', abbreviated as $r$(\textit{irf}, \textit{airfield}), this is to be understood as the correlation of the row vector of \textit{irf} in $\bm{F}$ and the row vector of \textit{airfield} in $\bm{S}$. Table~\ref{tab:airfield} shows that $r$(\textit{irf}, \textit{airfield}) is greater than both $r$(\textit{irf}, \textit{air}) and $r$(\textit{irf}, \textit{field}).  The same holds for the corresponding correlations of \textit{rfi}.   Furthermore, the trigrams that \textit{airfield} shares with its left constituent (\textit{\#ai, air}) are more strongly correlated with \textit{air} than with \textit{airfield}. Likewise, three of the trigrams that \textit{airfield} shares with its right constituent (\textit{fie, iel, ld\#}) are more strongly correlated with \textit{field} than with \textit{airfield}. The exception is \textit{eld}, which is negatively correlated with \textit{airfield}, and even more negatively correlated with \textit{field}.  To sum up, this example shows that the boundary trigrams support the meaning of the compound much better than the meanings of the constituents. Non-boundary trigrams tend to support the meanings of the constituents much better than they support the meaning of the compound itself.   This is unsurprising, as for instance \textit{air} has to be shared with many other compounds, in our dataset \textit{air-pump, air-raid, airbrake, aircraft,    aircrew, airfield, airflow, airforce, airframe, airgun, airlift, airline, airlock, airmail, airman, airplane, airport, airscrew, airship, airsick,    airspeed,  airstrip}, and \textit{airway}. 

\begin{figure}
    \centering
    \includegraphics[width=0.7\linewidth]{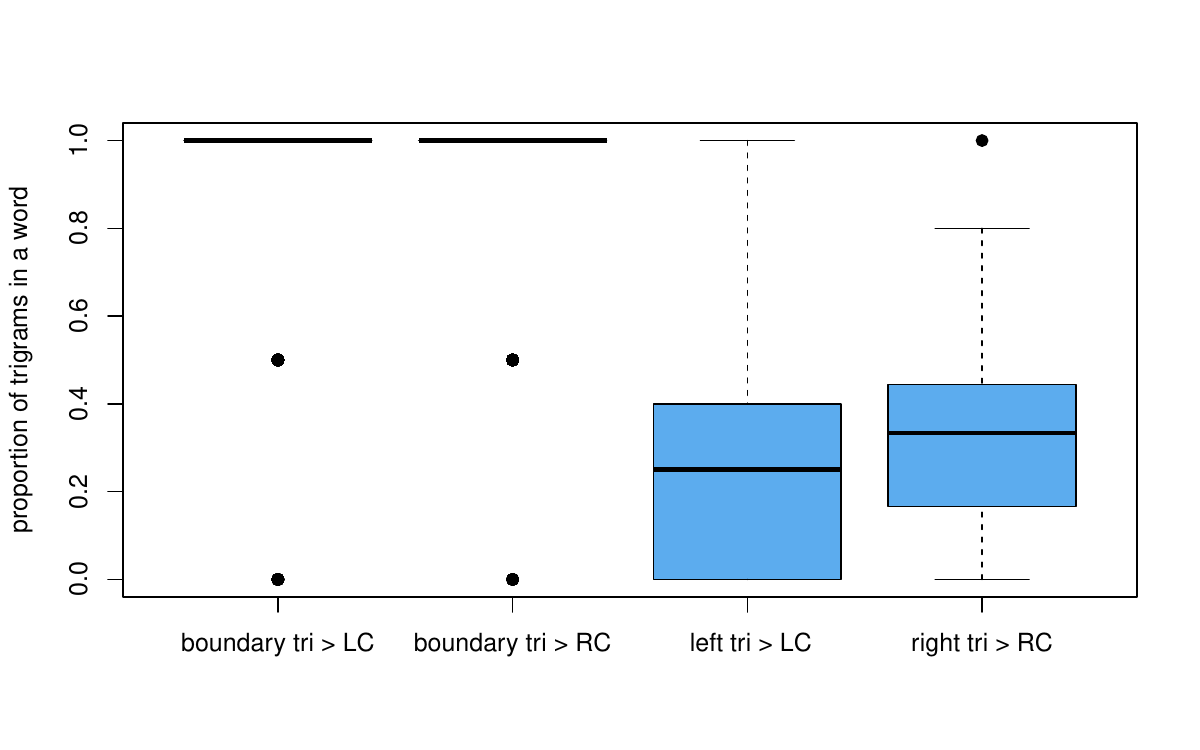}
    \caption{Proportion of trigrams in a word for which the correlation with the compound embedding exceeds the correlation with the embedding of a left (LC) or right (RC) constituent, broken down by boundary trigrams (boundary tri), trigrams preceding the boundary (left tri) and trigrams following the boundary (right tri).}
    \label{fig:boxplot_compounds}
\end{figure}

Figure~\ref{fig:boxplot_compounds} illustrates that this pattern of results is not unique to \textit{airfield}.  For the 2288 two-constituent compounds in our dataset for which a parse is available in the CELEX database \citep{Baayen:CD95}, this boxplot visualizes the distribution of proportions in a word of trigrams for which the correlation with the compound is greater than the correlation with the left or right constituent.  Trigrams are grouped into boundary trigrams, trigrams preceding the boundary (left tri), and trigrams following the boundary (right tri).  For some 83 to 84\% of the compounds, both trigrams have higher correlations with the compound embedding than with the left or right constituent embeddings. The proportions of compounds for which the trigrams are more correlated with the compound embedding than with the left or right constituent embedding are much lower.  In other words, it is the boundary trigrams that have the highest relative functional load of guiding interpretation to the meaning of the compound itself.

Now that we understand how the model manages to learn to understand compounds, it is not difficult to see why it cannot generalize very well. Boundary trigrams have to be shared with other compounds.  \textit{rfi} occurs not only in \textit{airfield}, but also in \textit{butterfingers, quarterfinal, silverfish, and starfish}. Likewise, \textit{irf} is also a boundary trigram for  \textit{airflow, airforce} and \textit{airframe}. As \textit{butterfingers, silverfish, airflow}, and \textit{airforce} are semantically highly diverse, \textit{irf} is not very useful for semantic generalization.

Limited generalization for compound semantics is possible, as shown by the CAOSS model of \citet{marelli2017compounding} applied to 8147 English compounds.  This model takes the embeddings of the constituents as input, and uses a linear transformation to approximate the embedding of the compound (see the appendix for technical details). Accuracy of this model for all compounds jointly is at 54.3\%, but in a training-testing set-up with  random selection of 815 held-out compounds, accuracy for the held-out data is down to 23.8\%.  Interestingly, just adding up the embeddings of the constituents for the held-out data creates predicted compound vectors that already have an accuracy of 18.0\%. The fine-tuning that the linear transformation of CAOSS offers over and above vector addition is an increase in accuracy of only 5.8\%. 

The low generalization accuracy of CAOSS fits well with the findings of \citet{schafer2020constituent}, who observed that native speakers of English have great difficulties guessing the meaning of novel compounds such as \textit{acid cap}, even when some of the original context of use is provided.  In the light of these considerations, it makes sense that the comprehension mappings of the DLM cannot predict the meanings of held-out compounds with precision.

When modeling the productivity of compounds in production, we again find that the training data can be learned well, but that generalization is highly inaccurate. This is illustrated by a mapping from 300-dimensional embeddings to the 4799-dimensional trigram vectors of the dataset of 11,170 English words. As mapping with precision from a low-dimensional space to a high dimensional space is usually impossible with a linear transformation, we made use of a deep network.\footnote{This network had 1000 RELU units, using the binary cross-entropy loss. For details of this kind of network, see \citet{Heitmeier:Chuang:Baayen:2024}.}  Using the same training-test split we used above, we found that for training data, this model predicted the form vector correctly for 92\% of the words.  However, for the held-out compounds, accuracy plummeted to 24.7\%.  The idiosyncratic meanings of the held-out compounds stand in the way of predicting their pronunciations.\footnote{As the present dataset does not include compounds written with internal spaces,  these  potentially more semantically transparent compounds \citep{Kuperman:Bertram:2013} (if these are not to be understood as phrases) are not taken into consideration in the present study.}
This brings us back to our original question: compounds are evidently productive type-wise, but how can this be given that there is no overarching systematicity enabling generalization?  We think that compounding is a highly useful for onomasiological tinkering, for creating names by reusing old words in creative ways. This `bricolage' \citep{strauss1962savage} tends to be locally motivated at time of creation (cf. the etymologies of opaque compounds such as \textit{dumbbell, moonshine} and \textit{hogwash}), and can lead to small islands of reliability in semantic space. However, the productivity of such islands of reliability may be severely constrained, as shown by \citet{shen2022productivity}. 

\citet{shen2022productivity} studied compounding in Mandarin Chinese, focusing on how often a given word appears as a constituent of a compound in a given position. Following their terminology, we refer to a position-specific constituent word as a `pivot'.  Examples of compounds with the pivot {\em da4}, `big' are {\em da4jia1} `everyone';  {\em da4xue2} `university' and {\em da4liang4} `generous'.  This study observed that as the number of compound types with a given pivot $V$ increases,  the Good-Turing probability of unseen types ${\cal P}$ \citep[the category-conditioned productivity measure proposed in][]{Baayen:92:YoM} decreases. Such a negative correlation of $V$ and ${\cal P}$ is absent for English derivational affixes.  \citet{shen2022productivity} also observed that as ${\cal P}$ increases, the embedding of the pivot is more similar to the embedding of its compound, as well as to the embedding of the non-pivot in its compound.  Considered jointly, these results suggest that semantic transparency is possible for pivots with small numbers of compounds, but that if a pivot is re-used in more and more compounds, semantic transparency suffers.

Interestingly, the productivity of a pivot decreases if the region in semantic space defined by the compounds of the pivot overlaps too much with the region in semantic space defined by the compounds of the non-pivot.  To define a semantic island of reliability for the compounds with a given pivot, they took the embeddings of these compounds, and calculated the centroid, the most prototypical meaning of the pivot in a compound. Next, they calculated, for each compound with the pivot, its correlation with the centroid.  For the resulting distribution of correlations, they calculated the 95\% confidence interval. All compounds having a correlation within this confidence interval around the centroid are considered to be in the pivot's island of reliability. Finally, \citet{shen2022productivity} calculated for all compounds with the non-pivot as constituent how many had a correlation with the pivot's centroid 95\% confidence interval.  They designated such non-pivot compounds as `intruders'.   The count of intruders was shown to be negatively correlated with ${\cal P}$.  Furthermore, the average of the correlations of the embedding of the pivot with the embeddings of its compounds, a measure of the transparency of the pivot, is negatively correlated with the number of intruders.  

In our study of Finnish nominal inflection, the centroids of case-number combinations emerged as a good approximation of the contribution that in the DLM word final trigrams (which capture a substantial part of the exponents for case and number) make to the predicted embedding of the inflected noun.  For compounding, our conjecture is that the centroids of pivot constituents are also important, and the more semantically related words are within the area around the centroid, the more productive the pivot can be.  However, a pivot that is originally semantically motivated can be put to other uses that move away from the centroid. This  onomasiological tinkering rides piggy back on the popularity of a pivot, but because human-perceived order in the natural world is highly constrained by the diversity of the natural world and the limitations of human perception \citep{kant1999critique,Husserl:1913,merleau2013phenomenology,hoffman2019case}, for productivity, the more tinkering goes on at time $t$, the less tinkering is attractive at time $t+1$. An additional complication is that compounds with the non-pivot can enter the pivot's island of reliability and mess up its semantic transparency. 

In summary, morphological tinkering is great. Tinkering poses no problem for learning of training data.  L1 learners likely don't face much of a problem, in part because subliminal statistical learning is easiest when young, and in part because vocabulary development goes hand in hand with general broadening of cognitive skills.  However, we should not expect compound tinkering to afford much generalization for test data. 
L1 learners can make use of contextual inferencing skills, but specifically L2 learners face a formidable memorization task, as they are more likely to attempt decompositional strategies which run aground as trying to predict what novel compounds mean is largely self-defeating.  At the micro-level of individual pivots some generalization is possible, perhaps, but not at the macro-level.  \citet{Schultink:62} argued that compounds do not constitute a morphological category. We agree: compounds are sui generis, and very useful and productive as onomasiological device, albeit without relying (much) on form-meaning generalization.  As a consequence, systemic productivity is low: our estimate for English production is 24.7\%,  and our estimates for comprehension are 12.6\% for endstate learning, and 0 for frequency informed learning.

In the next section, having completed our examination of the extremes of inflection and compounding, we turn to derivation.

\subsection{Derivation}

A language that is particularly rich in morphological derivation is Malay, an Austronesian language spoken by more than 200 million peoples in various countries of Southeast Asia. Malay has minimal inflection. Affixation, by means of derivational prefixes, is at the heart of what makes Malay productive. In Malay, allomorphy is pervasive such as in prefix families of \textit{beR-} (i.e., be-, bel-, and ber-), \textit{meN-} (i.e., me-, men-, mem-, meny-, meng-, and menge-), \textit{peN-} (i.e., pe-, pen-, pem- peny-, peng-, and penge-), \textit{peR-} (i.e., pel-, and per-), and \textit{teR-} (i.e., te-, and teR-), some of which typically change the meaning or class of a word. In some cases, the initial letter of the stem is omitted to facilitate pronunciation (e.g., \textit{fikir}/think; \textit{\textbf{pem}ikir}/a thinker). Although this allomorphy complicates systematicities between form and meaning, \citet{mohamed2025exploratory} have shown that, nevertheless, derived words cluster by prefix in semantic space. In that study, a \textit{t}-SNE analysis (\citealp{maaten2008visualizing}) was conducted on a large set of complex Malay words using high-dimensional word embeddings. Following a key principle of Distributional Semantics, the authors observed a strong correspondence between form and meaning such that words that share a prefix appear closer in semantic space than those that contain a different prefix. 

From a computational perspective, generalization is as an index of productivity. A question of primary interest concerns how well generalization occurs in a language in which morphological derivation is extensive. To address this question, the approach taken is two-fold. First, as demonstrated in our working examples for inflection and compounding, we trained the DLM on a set of Malay words, and evaluated its comprehension accuracy on words in the training set and test accuracy on the held-out data. The accuracy with which the DLM predicts the meaning of the unseen forms in the held-out data is contingent on the statistical co-occurrences between form and meaning of the words in the training set. From the series of analyses above, regularities in form and meaning appear rather transparent for inflection and very opaque for compounds. If our intuition holds, the DLM trained on Malay derived words should generalize to held-out data much better than English compounds, but pale in comparison to Finnish inflected forms. Because the DLM is also a cognitive model, a secondary goal is to assess its performance against behavioural data. Implications for the ease with which we read are discussed as a function of form-meaning systematicities.

Our dataset comprises of 8,843 words from the Malay Lexicon Project \citep{yap2010malay} for which FastText embeddings \citep{bojanowski2017enriching} were extracted. Of these words, 7,959 words make up the training set and the remaining 884 words were reserved as held-out data for testing, such that all cues and derivational features present in the test set appeared in the training set (for details on careful splitting, see \citet{heitmeier2024discriminative}). Words' forms were represented using 4-gram vectors and words' meanings were represented using FastText embeddings. For both end-state learning and frequency-informed learning, see Table~\ref{table:dlm_acc_malay} for the overall comprehension accuracies for the training and test data. Accuracies were evaluated at the top one and top 10 percent, that is, whether the predicted vector was the closest or among the top 10 closest neighbors of the targeted gold-standard embedding. 

Indeed, the learning and test performance of the DLM trained on Malay derived words was somewhere in between the accuracies of the models trained on Finnish inflected words and English compounds. Most obvious is the large discrepancy in test accuracy for Malay derived words compared to English compounds, in that generalization to novel forms for derived words far exceeded that of compounds. In contrast, the DLM outperformed in both learning and test performance when the model was trained on Finnish inflected words than on Malay derived words. It is clear that the extent of form-meaning regularities of the data on which the model was trained on closely corresponds to the rate at which successful generalization occurs. In sum, these comparisons further iterate the need for a strong correspondence between form and meaning for generalizations to be successful. 

\begin{table}[ht]
    \centering
    \caption{Comprehension accuracies for training and test data (Types).}
    \label{tab:derivational_accuracies}
\begin{tabular}{lrrrr} \hline 
           & \multicolumn{2}{c}{training} & \multicolumn{2}{c}{testing} \\ \hline 
  learning &  @1     & @10    &  @1    &   @10 \\ \hline
  EOL      &  0.870  & 0.907  & 0.289 & 0.594 \\
  FIL      &  0.341  & 0.427  & 0.146      & 0.370 \\ \hline  
\end{tabular}
\label{table:dlm_acc_malay}
\vspace{0.5em} 
    \begin{minipage}{\textwidth}
        \centering
        \small
        \textit{Note.} For the training data, token-wise accuracy @1 for FIL is 0.90.
    \end{minipage}
\end{table}

\begin{figure}[H]
    \centering
    \includegraphics[width=0.4\linewidth]{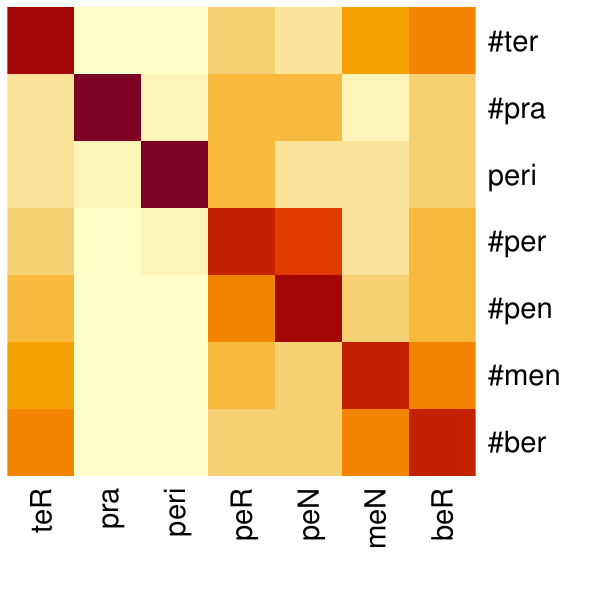}
    \caption{Correlation heatmap of prefix centroid embeddings and a subset of the row vectors of the F matrix that maps the form vectors (4-gram) to its meaning (FastText embedding) in the DLM. The set of 4-grams presented correspond to at least one of the prefixes. Darker hues, compared to lighter hues, indicate a stronger correlation between the embeddings of the centroid and row vectors of the F matrix. The strongest correlations are present for the 4-grams that overlap most with the prefix, indicating that it is the prefixal 4-grams that contribute most to realizing the meaning of the centroid.}
    \label{fig:fourgram_hm}
\end{figure}

To help us better understand how the model learns derived words, as illustrated using compounds, we extracted the comprehension F mapping, obtained with end-state learning. Here, we correlated a small subset of the row vectors of F with the embeddings of the prefix centroids, that is, the mean embeddings of words containing a particular prefix. \citet{mohamed2025exploratory} provided empirical evidence for Malay that the correlation between the embedding of a derived word and the centroid was a reliable measure of semantic transparency (i.e., form-meaning systematicity) and the best predictor of lexical decision latencies, among several simple measures of transparency. A stronger correlation with the centroid indicates a more consistent mapping between form and meaning. Because the F mapping is a very large matrix, the row vectors we first analyzed were constrained to word-initial 4-grams that share a spelling with at least one of the prefixes (e.g., \textit{\#ber-}; prefix \textit{beR-}). We tested whether the semantic contribution of the row vectors of prefix-like initial n-grams in the F matrix were similar to the centroids of the corresponding prefixes. Figure~\ref{fig:fourgram_hm} depicts that that is the case. Not shown here, using 3-grams instead of 4-grams, the same analyses on the F matrix obtained from the DLM trained on the same set of words yield similar results. Then, we further examined the top five 4-grams in the F mapping for which embeddings are most correlated with each centroid. For most prefixes, the prefixal 4-grams contribute most strongly to the meaning of each centroid (\textit{r} \texttt{>} .80; see Table~\ref{tab:top5_ngram_malay}). Prefix-like 4-grams are only moderately correlated with the centroids of \textit{peri-} and \textit{pra}. \textit{Peri-} occurs in very few words in our dataset and appears somewhat more opaque in its mapping between form and meaning (e.g., \textit{peribahasa}/figure of speech, \textit{perihal}/the state of something; \textit{perilaku}/one's behaviour). On the other hand, \textit{pra-} appears rather transparent in its form-meaning correspondence, and is semantically similar to the English \textit{pre-}, such as in \textit{prauniviersiti}/preuniversity. \textit{Pra-}, however, is loaned from Sanskrit and also does not occur frequently in Malay. It is also noteworthy to point out that the row vectors of stem and word-final 4-grams are moderately correlated with each of the centroids as well. In particular, 4-grams that include the first segment of the stem are correlated with prefixes that have fewer than four letters (e.g., \textit{pra}s). Several word-final 4-grams that are correlated with the prefix centroids likely include the suffix \textit{-an}, such as in \textit{aan\#}, \textit{san\#}, and \textit{uan\#}. \textit{-an} is highly productive and occurs in more than 60 percent of words containing a suffix in our dataset. Such inspections on the intercorrelations among word-initial, stem, and word-final 4-grams with the centroids provide clarity as to why generalization occurs more seamlessly for derived words relative to compounds, but less extensively compared to inflected forms.

\begin{table}[H]
\centering
\caption{
Top five strongest correlations between the row vectors in $\bm{F}$ and the centroid embeddings of selected prefixes. Coefficients highlighted in red represent the strongest correlation between the row vectors of the 4-gram and the embeddings of the centroid. Each prefix, except \textit{peri-}, are most correlated with at least one of their allomorphs, providing evidence of differences in degrees of productivity for allomorphs.
}
\label{tab:top5_ngram_malay}
{\footnotesize
\begin{tabular}{r|rrrrrrrrrr} \hline
 & beR & meN & peN & peR & peri & pra & teR \\ \hline
  \#ber  & {\color{red}\textbf{.847}} & .717 & - & - & - & - & .701\\ 
  itu\# & .739 & .720 & .677 & .682 & - & - & .752\\ 
  aan\# & .686 & - & .755 & .802 & .572 & .588 & .669\\ 
  \#per & .672 & - & .768 & {\color{red}\textbf{.809}} & {\color{red}\textbf{.580}} & .549 & - \\ 
  bat\# & .668 & - & - & - & - & - & - \\ 
  \#men & - & {\color{red}\textbf{.839}} & - & - & - & - & -\\ 
  \#mem & - & .753 & - & - & - & - & -\\ 
  \#mel & - & .740 & - & - & - & - & -\\ 
  \#pen & - & - & {\color{red}\textbf{.803}} & - & - & - & -\\
  \#pem & -  & - & .670 & - & - & - & -\\
  san\# & -  & - & - & .657 & - & - & -\\
  uan\# & -  & - & - & .653 & - & - & -\\
  \#kes & -  & - & - & - & .498 & - & -\\
  \#kei & -  & - & - & - & .497 & - & -\\
  \#kea & -  & - & - & - & .478 & - & -\\
  \#pra & -  & - & - & - & - & {\color{red}\textbf{.684}} & -\\
  pras & -  & - & - & - & - & .574 & -\\
  gka\# & -  & - & - & - & - & .516 & -\\
  \#ter & -  & - & - & - & - & - & {\color{red}\textbf{.822}}\\
  lah\# & -  & - & - & - & - & - & .654\\
  \hline
\end{tabular}
}
\end{table}

From a psycholinguistic perspective, a related question concerns how regularities in form and meaning (and, by extension, productivity) influence reading comprehension. To address this matter, we used two measures extracted from the DLM to predict response latencies. These measures are \textit{Target Correlation} and \textit{r target}. In principle, \textit{Target Correlation} and \textit{r target} are conceptually similar. Both measures refer to the correlation between the semantic vectors of the predicted word and the target, except that the estimates from \textit{Target Correlation} corresponds to words in the training set, and estimates from \textit{r target} corresponds to words in the held-out data. In the subsequent text, we refer to them as \texttt{Target Correlation$_{\text{train}}$} and \texttt{Target Correlation$_{\text{test}}$} respectively. Lexical decision latencies from a series of previous studies of the Malay Lexicon Project \citep{maziyah2023malay,maziyah2023distributional,maziyah2025malay}, were extracted for 1,624 words in the training set and for 200 words in the held-out data. In what follows, we report our results on two sets of GAMs with careful consideration of concurvity statistics.

\begin{figure}[H]
    \centering
    \includegraphics[width=0.8\linewidth]{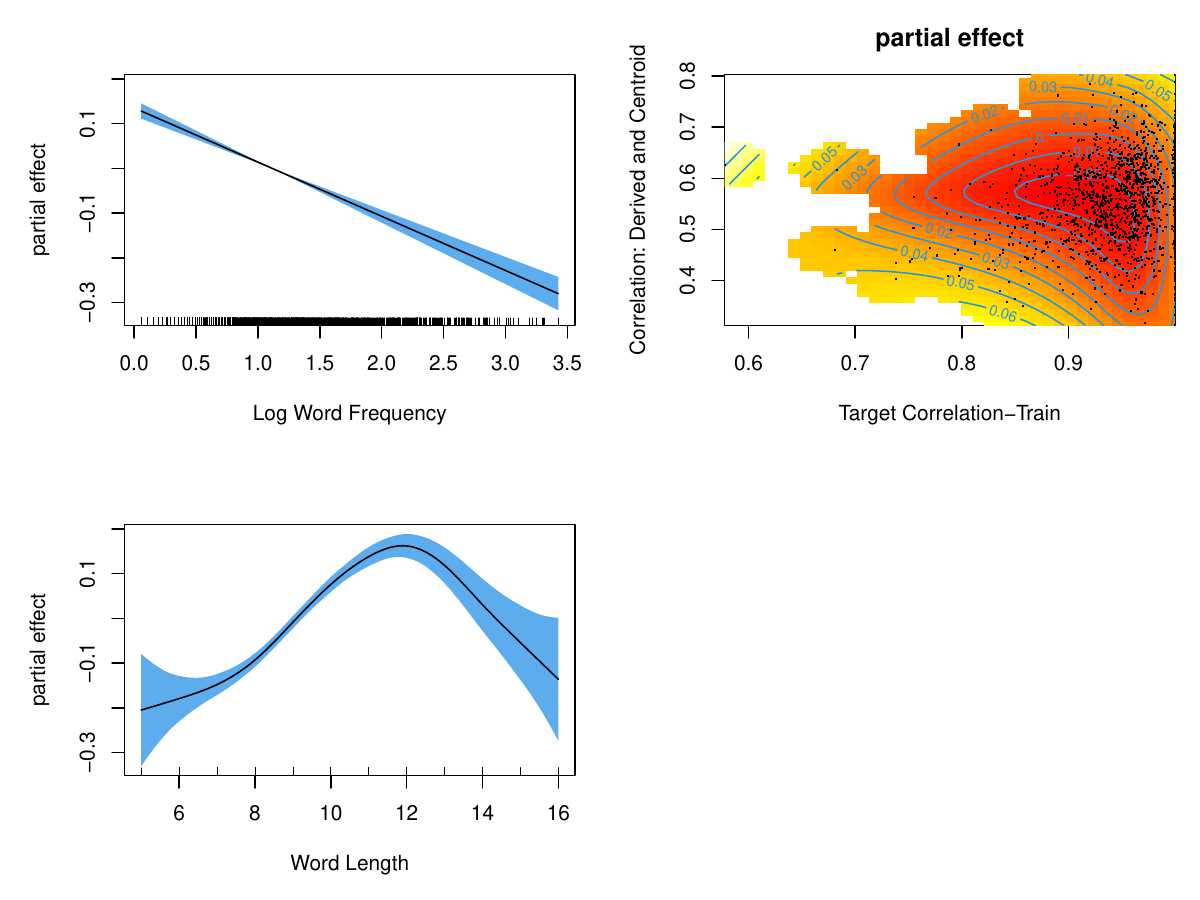}
    \caption{Partial effects of the GAMM fitted to the training data. Left column: Partial effect of word frequency on RT (top). Partial effect of word length on RT (bottom). Rugged lines on the x-axes represent the data. Right column: Partial effect of the interaction between \texttt{Target Correlation$_{\text{train}}$} of the DLM and the correlation between each derived word and their centroid. Red hues denote shorter RTs, orange/yellow hues indicate longer RTs. Black dots represent the data.}
    \label{fig:datbam_training}
\end{figure}

First, we fitted a GAM to the training data. Since \texttt{Target Correlation$_{\text{train}}$} and \texttt{Target Correlation$_{\text{test}}$} were extracted from the DLM trained with end-state learning, we take into account word frequency and word length. Word frequency and word length are well-established predictors of word recognition. Word frequency was logarithmically transformed and RTs were inverse transformed. Of interest, \texttt{Target Correlation$_{\text{train}}$} and the correlation between the derived word and the centroid were entered as predictors of RT, with prefix as a random effect. All predictors significantly predicted RT (see Table~\ref{table:gamm_training}). The top left panel of Figure~\ref{fig:datbam_training} shows a facilitative effect of word frequency on RT, that is, faster responses were elicited for higher frequency words than for lower frequency words. The observed effect of frequency on RT is consistent with prior literature. In the bottom left panel of Figure~\ref{fig:datbam_training}, we observe an inhibitory effect of length, and a somewhat surprising facilitative effect for very long words. Note that very few words contained 13 or more letters (about 4 percent of the data). The effect of the correlation between the derived word and the centroid on RT mirrored recent findings in Malay \citep{mohamed2025exploratory}. In that study, the effect of the correlation between derived words and their centroid embeddings on RT appeared non-linear and U-shaped. More specifically, faster responses were first observed the stronger the correlation between the embeddings of derived words and those of their centroids, followed by slower responses for very strong correlations (\textit{r}\texttt{>}.7). Similarly, here, in the top right panel of Figure~\ref{fig:datbam_training}, responses were faster the closer a word is to its centroid, with the fastest responses predicted for words that share a moderate correlation with their centroid. However, responses were much slower for words that very closely resembled their centroid.

\begin{figure}[H]
    \centering
    \includegraphics[width=0.8\linewidth]{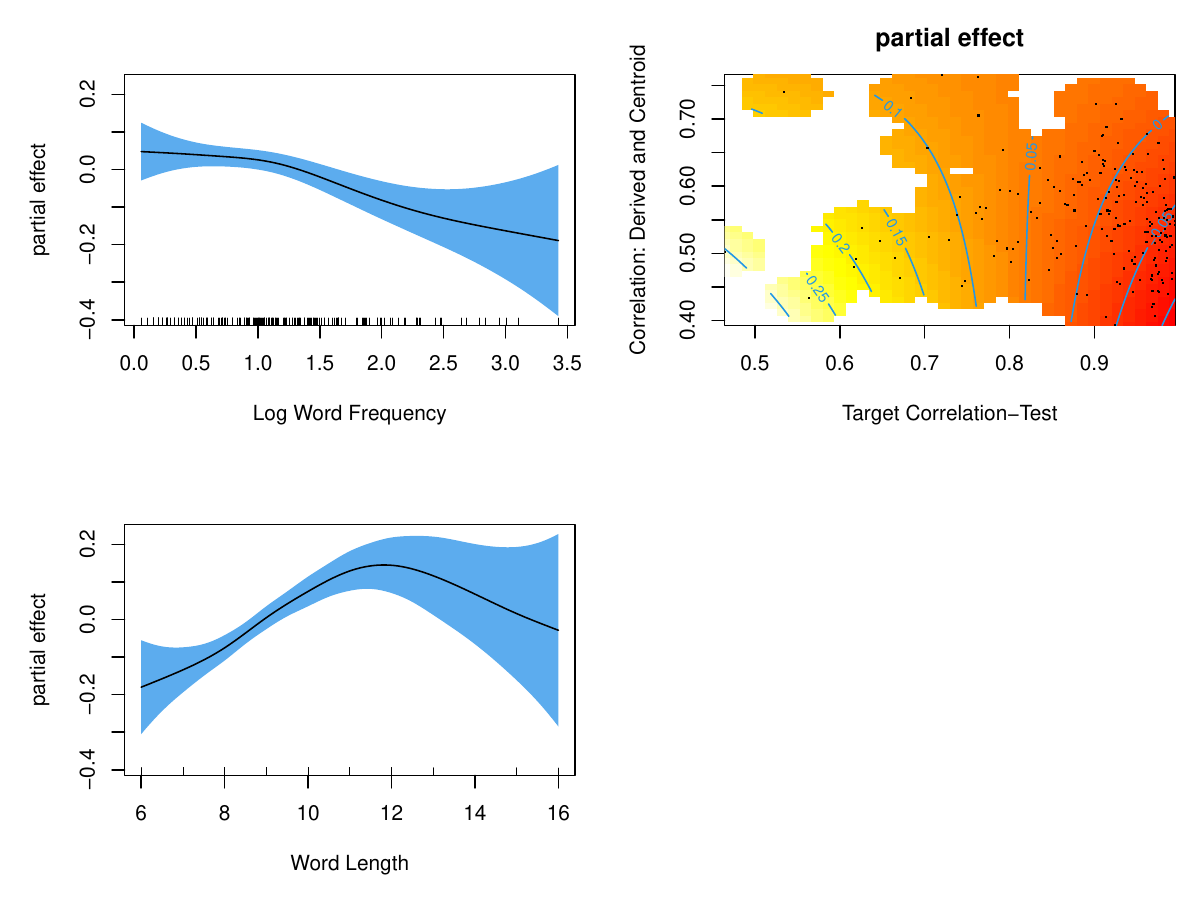}
    \caption{Left column: Partial effect of word frequency on RT (top). Partial effect of word length on RT (bottom). Rugged lines on the x-axes represent the data. Right column: Partial effect of the interaction between \texttt{Target Correlation$_{\text{test}}$} and the correlation between each derived word and their centroid. Red hues denote shorter RTs, orange/yellow hues indicate longer RTs. Black dots represent the data.}
    \label{fig:datbam_test}
\end{figure}

A separate GAM was fitted only to the held-out data. As in the GAM analysis above, the same predictors were entered in the model, except that \texttt{Target Correlation$_{\text{test}}$} was entered as a predictor in place of \texttt{Target Correlation$_{\text{train}}$}. Again, all predictors significantly predicted RT (see Table~\ref{table:gamm_test}). Each panel of Figure~\ref{fig:datbam_test} shows that the effect of each predictor in the analysis of held-out data exhibited similar trends on RT as illustrated above with the training data.

\begin{table}[H]
    \centering
    \caption{GAMM fitted to the training data.
    Word frequency was log-transformed. TargetCorTrain = \texttt{Target Correlation$_{\text{train}}$}. CorDev-Centroid = Correlation between derived words and their centroid. The model syntax is inverse RT $\sim~$ s(log frequency) + te(TargetCorTrain*Corr.Dev-Centroid + s(word length) + s(prefix, bs= ‘re’). Inverse RT = -1000/RT; a negative sign is used to ensure that the transformed RT and the observed RTs are positively correlated, facilitating interpretation. 
    }
        \resizebox{\textwidth}{!}{
            \begin{tabular*}{\textwidth}{@{\extracolsep\fill}lrrrrrr}
            \hline
            {\bf Parametric coefficients} \\
            \hline
            Variable & Estimate  & Std. Error & {\it t}  &  {\it p} \\
            \hline
            Intercept & -1.10   & .04  & -29.00 & \texttt{<}.0001\\
            \hline
            {\bf Smooth terms}\\
            \hline
            Variable & edf  & Ref.df & F  &  {\it p}\\
            \hline
            Frequency  & 1.00  & 1.00  & 225.94 & \texttt{<}.0001\\
            TargetCorTrain*CorDev-Centroid  & 7.78   & 9.87   & 1.88 & .0451\\
            Word length & 5.34  & 6.40  & 57.95 & \texttt{<}.0001\\
            Prefix      & 7.64 & 9.00 & 37.78   & \texttt{<}.0001\\
            \hline
            $R^2$ = .464\\
            \hline
            \end{tabular*}
}
\label{table:gamm_training}
\end{table}

\newpage

\begin{table}[H]
    \centering
    \caption{GAMM fitted to the held-out test data.  Word frequency was log-transformed. TargetCorTest = \texttt{Target Correlation$_{\text{test}}$}. CorDev-Centroid = Correlation between derived words and their centroid. The model syntax is inverse RT $\sim~$ s(log frequency) + te(TargetCorTest*Corr.Dev-Centroid + s(word length) + s(prefix, bs= `re'). Inverse RT = -1000/RT.}
        \resizebox{\textwidth}{!}{
            \begin{tabular*}{\textwidth}{@{\extracolsep\fill}lrrrrrr}
            \hline
            {\bf Parametric coefficients} \\
            \hline
            Variable & Estimate  & Std. Error & {\it t}  &  {\it p} \\
            \hline
            Intercept & -1.14   & .04  & -25.84 & \texttt{<}.0001\\
            \hline
            {\bf Smooth terms}\\
            \hline
            Variable & edf  & Ref.df & F  &  {\it p}\\
            \hline
            Frequency  & 2.26  & 2.83  & 4.44 & .00527\\
            TargetCorTest*CorDev-Centroid  & 3.00   & 3.00   & 4.29 & .00590\\
            Word length & 3.21  & 4.02  & 8.27 & \texttt{<}.0001\\
            Prefix      & 5.04 & 8.00 & 4.40   & \texttt{<}.0001\\
            \hline
            $R^2$ = .401\\
            \hline
            \end{tabular*}
}
   \label{table:gamm_test}
\end{table}

\clearpage

Ideally, the effect of a productive system on cognition is facilitative for reading comprehension, allowing more automatic processing. Our findings on Malay provide evidence that this is largely the case, albeit not entirely without trade-offs. A very productive system also implies a densely populated morphological space. A word that highly resembles the prototypical meaning of many morphologically related words imposes additional demands. Not only must the reader navigate through large and tightly clustered word representations, but also expend additional cognitive resources in discriminating between related words.

\subsection{Discussion}

We have shown that generalization to unseen data, the hallmark of systemic morphological productivity, depends on systematic correspondences between form and meaning.  Even in a complex semi-agglutinative system as that of Finnish, the DLM model capitalizes on the systematic correspondences between form and meaning for combinations of case and number.  For prefixation in Malay, form-meaning correspondences are also detecable, although here the functional load of n-grams (as the fuzzy counterpart of morphemes and allomorphs) is shared with both n-grams of the stem and n-grams of suffixal exponents.  For compounding, we have argued that tinkering is the modus operandi, with micro-systematicities at the level of individual pivots playing a minor role. Apparently, we have to distinguish between two fundamentally different kinds of productivity: systemic productivity on the one hand, and the productivity of bricolage on the other hand. 

And yet, all the simulations above are flawed. Perhaps not deeply flawed, but flawed they are. The reason is that the DLM is a cognitive model of an individual speaker, but we have been using corpus data, i.e., aggregate data from large communities of speakers.  Within such communities, language users have different areas of expertise, each of which come with their own specialized vocabularies.  No single speaker is an expert on the vocabularies of all socio-cultural expertise (lexicographers being a potential counterexample).  Testing on held-out data is likely unfair for any individual speaker.  
In the next section, we present an attempt to address this discrepancy between community behavior (i.e., corpus data) and a computational model for an individual brain (the DLM).

\section{Productivity of a concrete individual in their socio-historical context}

So far, all measures and models were on the level of the language community. But all language users have their unique window onto language use, their own tailored input, which in turn shapes the way they use language. No two people receive exactly the same utterances. This basic insight of usage-based models makes it necessary to also take the level of the individual language user into account. Ideally, we need a compilation of everything one specific person has ever heard and read, and a compilation of everything they have ever said or written. For a variety of reasons, this is utopian (or rather, dystopian). 

We can curate, however, something similar for prolific authors who also kept journals and/or regularly kept correspondences. For these persons, it is in principle possible to reconstruct their reading diet from their journals and letters, and to also collect their output from published texts, manuscripts, journals, letters, etc. This is what we have started to do with German writer Thomas Mann (1875-1955), the 1929 Nobel Prize in Literature laureate. One may argue that our focus on written input introduces a bias into our data. That is certainly true. However, Mann's spoken input is beyond reach, and most of the new words and syntactic constructions he encountered he probably encountered through reading.

\subsection{Data}
We reconstructed Mann's reading diet on the basis of his diaries (1918-21, 1933-1955), his letters \citep{bürginmayer1977} and his personal library in Zurich. Each text was then collected; for practical reasons, we limited our search to texts that are digitally available on the internet. There are four classes of sources that vary in their quality:
\begin{itemize}
    \item edited digitized versions from literature hubs (``Projekt Gutenberg'', ``zeno.org'', ``Wikisource'') --- excellent quality;
\item scanned versions with text extracted with optical character recognition (OCR) from university websites and Google Books --- acceptable quality;
\item digitized versions from other sites (e.g., Internet Archive), OCR'ed with Tesseract --- mostly inadequate quality;
\item scans from Mann's personal library in Zurich's website (\url{https://nb-web.tma.ethz.ch/}) --- single page scans of texts which were only marginally used due to excessive effort.
\end{itemize}

So far, we have collected information on Mann's readings up until the year 1925. Of these, texts until and including 1915 have been collected. Our input corpus consists of 437 texts that Mann states to have read until then, containing 41.4 million tokens. The output corpus consists of 32 texts that Mann wrote during that time, and it contains 500,000 tokens. These numbers show a striking imbalance between Mann's readings and his writings: Mann read at least 80 times more than he wrote.

The Mann corpus is work in progress and will be expanded continually. The results presented here should be seen as a proof of concept that such a text collection is indeed useful and enables new insights. With that said, there are several shortcomings that have to be noted:
\begin{itemize}
    \item The first documented texts in the input corpus are from 1889, when Mann was 14 years old; there are hardly any records of his readings in earlier years. His adolescent reading diet is mostly terra incognita. 
    \item Newspapers have not been digitized yet. While we know which papers Mann regularly read, it is impossible to know which articles he actually read in which issue.
    \item Only a fraction of Mann's letters has been added to the corpora so far. While the letters he wrote are conveniently collected in edited volumes, the letters he received are often scattered, which makes addition a time-consuming task.
    \item Some OCR text files are considerably worse than the rest (e.g., "Geschichte der Lustseuche, die zu Ende des 15. Jahrhunderts in Europa ausbrach" by Philipp Gabriel Hensler from 1805). These texts are excluded for the construction of word embeddings (see below).
\end{itemize}

However, even if the corpus is not a complete collection of everything Mann ever read (and never will be, for principled reasons), it is possible to use it to investigate his word formations. The corpus may be skewed, but if it is, it is skewed towards the output: We systematically overestimate the number of newly coined formations. 

\begin{figure}[h]
\centering
\includegraphics[width=0.75\linewidth]{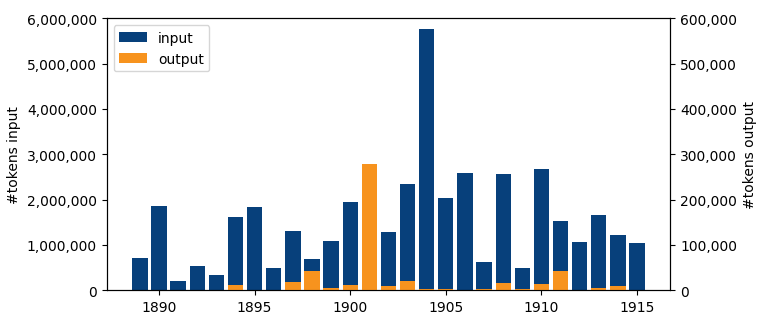}
\caption{\label{fig:running_words}Distribution of running words in the input and the output part of the Mann corpus. The number of words that Mann read exceeds the number of words he wrote by a factor of around 80.}
\end{figure}

Figure~\ref{fig:running_words} shows the distribution of input texts and output texts (note the different scales for the input texts, left side, and the output text, right side).

We investigate seven derivational patterns in the Mann corpus, four adjectival and three nominal ones. In previous research \citep{schneider2011produktivitat}, \textit{-los}, \textit{-bar}, and \textit{-tum} have been judged to be productive derivational patterns at the beginning of the 20th century (using \citet{Baayen:92a}'  potential productivity measure), while \textit{-nis} is unproductive \citep[121f.]{schneider2011produktivitat}. The pattern \textit{-sal}, which \citet{schneider2011produktivitat} did not investigate, was probably unproductive as well \citep{fleischer2012wortbildung}. \textit{-lich} and \textit{-sam} are both more or less unproductive hapax-wise, but \textit{-lich} exhibits around seven times as many types as \textit{-sam} in a large diachronic corpus of German \citep[141f.]{schneider2011produktivitat}. \textit{-bar} and particularly \textit{-los} are productive \citep[141f.]{schneider2011produktivitat}. Table~\ref{table:patterns_measures} gathers the relevant information on the different patterns.

\begin{table}[h]
\centering
\begin{tabular}{l|c|c|c|c}
Pattern & POS & example & potential productivity & \#V \\\hline
-bar & A & trinkbar 'drinkable' & 0.07 & 274\\
-los & A & herzlos 'heartless' & 0.11 & 190\\
-lich & A & freiheitlich & 0.02 & 354 \\
-sam & A & arbeitsam & 0.02 & 54\\\hline
-tum & N & bürgertum 'bourgeoisie' & 0.06 & 39 \\
-nis & N & ergebnis 'result' & 0.004 & 45 \\
-sal & N & trübsal 'misery' & -? & -?
\end{tabular}
\caption{\label{tab:productivity}Patterns of interest and their supposed potential productivity values P (from Schneider-Wiejowski, 2011: 108, 114, 129, 132).}
\label{table:patterns_measures}
\end{table}

Note that these are measures collected at the level of the language community. In the present paper, we will investigate whether these productivity values are mirrored at the level of an individual language user. It will also be curious to check the case of \textit{-lich}, which is an outlier: an unproductive pattern with a large number of attested types.

To study the semantic side of word-formation products, we constructed word embeddings.  All raw text files were lemmatized and PoS-tagged using the ParZu dependency parser for German \citep{sennrich2009new}, one of the most accurate parsers available \citep{ortmann2019evaluating}. All lemmas were checked against a large lemma database of contemporary German which contains 326,946 entries \citep{derewo2013vww} to automatically identify texts with a large degree of misspellings and errors due to faulty OCR. These misspellings and errors introduce unwanted variation when constructing word embeddings. We excluded the bottom quartile of texts ($> 60$\% of types not recognized from the DeReWo list, 107 texts) from further processing; this leaves us with 330 texts with 16.5 million tokens. From these, we excluded punctuation marks and words tagged as either proper nouns or foreign words. Following a suggestion in \citet{Baayen:Chuang:Shafei:Blevins:2019}, we split sentences longer than ten words into shorter clauses wherever possible, using the next conjunction, relative pronoun, interrogative pronoun, or punctuation mark as a break. The resulting sentences and clauses were then used to construct 300-dimensional Word2Vec embeddings \citep{mikolov2013distributed} from the texts' lemmas that span the whole time period of interest (parameters: window size = 5, iterations = 10, minimal count = 5, epochs = 15). The resulting semantic model consists of Mann-specific embeddings for 42,966 lexemes.

\subsection{Descriptive results}
Table~\ref{table:new_types_patterns} shows, for each pattern, the number of types and tokens in Mann's readings and writings, together with the number of hapax legomena in the input and, most importantly for our investigation, the number of new types in Mann's writings, words he has not read or written before.

\begin{table}[h]
\centering
\begin{tabular}{l|c|c|c|c|c|c}
 \hline 
           & \multicolumn{3}{c|}{input} 
           &  \multicolumn{3}{c}{output} \\ \hline 
Pattern & \#types & \#tokens  & \#hapaxes & \#types & \#tokens & \#new types \\\hline
\textit{-lich} & 991 & 300,476 & 2,661 & 351 & 4,922 & 17 \\
\textit{-los} & 676 & 22,215 & 2,082 &  140 & 511 & 11\\
\textit{-tum} & 335 & 11,169 & 510 & 31 & 95 & 9 \\
\textit{-bar} & 404 & 22,884 & 874 & 45 & 364 & 5\\
\textit{-sam} & 88 & 26,430 & 307 & 28 & 641 & 1\\
\textit{-nis} & 79 & 35,010& 172 & 41 & 452 & 0 \\
\textit{-sal} & 12 & 5,274& 43 & 5 & 49 & 0 \\

\end{tabular}
\caption{\label{table:new_types_patterns}Patterns of interest and their types, tokens, hapaxes in the input and output section of the Mann corpus}
\end{table}

The number of new types is surprisingly low. In total, the patterns in question have 43 new formations among them, over the course of 26 years. On average, that amounts to less than two new types per year. At least for these derivational patterns, Mann rarely goes beyond what he has read.

The number of types in the input and the number of new types in the output are highly correlated (Spearman's $\rho = 0.95, p = 0.0008$). The number of different types in the input is a good predictor for new types in the output. Note that the total number of types in the input (2,585) exceeds the number of new types in the output (43) by a factor of 60:1. In a way, then, these word formation patterns exhibit a rather leaky transmission.

The summed presentation in Table~\ref{table:new_types_patterns} gives a first overview, but it omits the temporal dimension that is also contained in the data. For each pattern, we can plot vocabulary growth curves \citep{Baayen:2001} for words in the input and in the output. For the output, we can further distinguish between words that Mann had encountered in his reading, and which left traces in his memory and may have been repeated, and those that are not contained in the input, and which he probably newly formed. 

\begin{figure}
    \includegraphics[scale=0.4]{"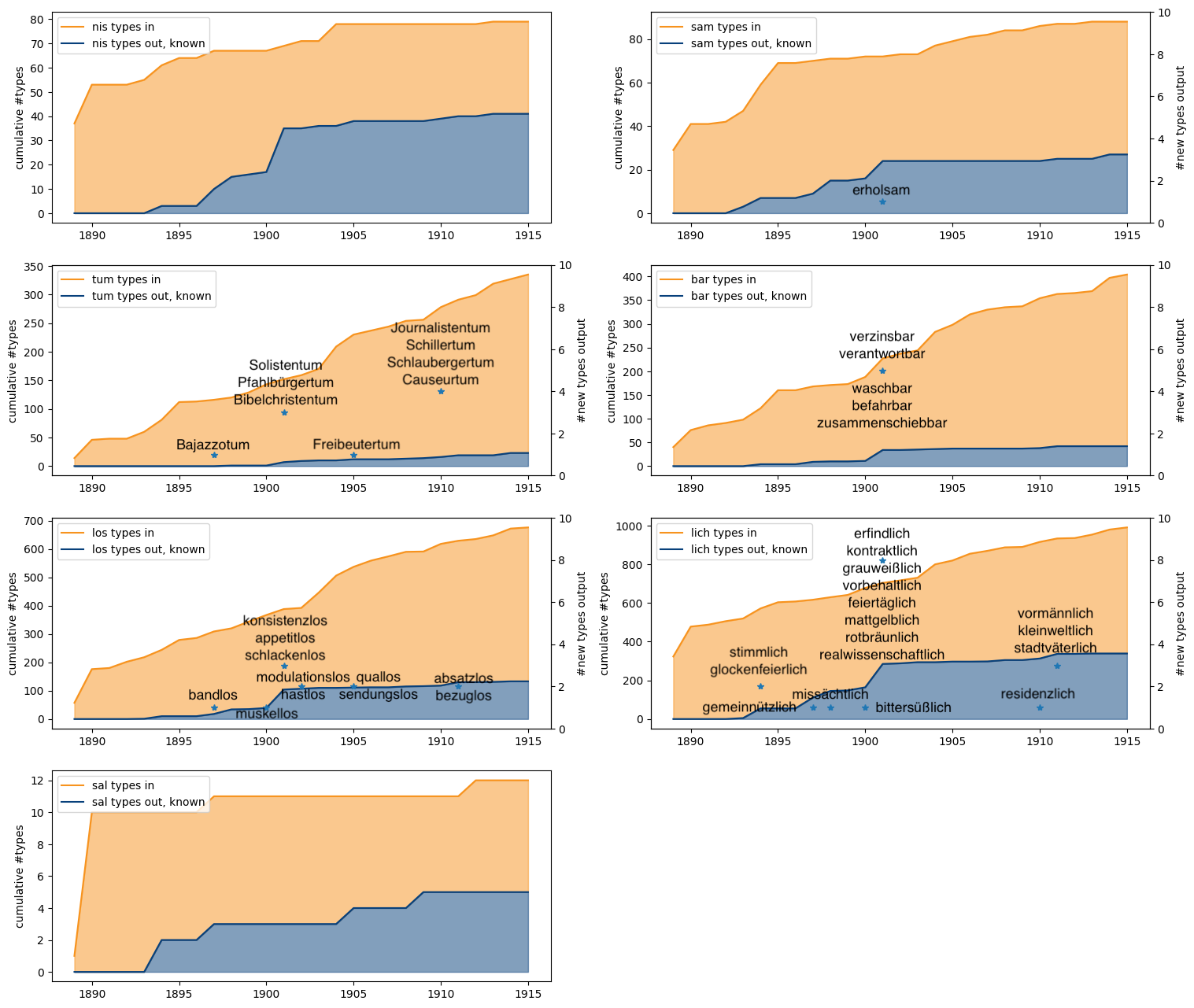"}
    \caption{\label{fig:pattern_descriptive_labels}Vocabulary growth curves for the input (orange) and the output for the seven patterns. Output growth curves are plotted for words known from the input; new types that exceed the input are plotted as asterisks (axis on the right hand side). }
    \label{fig:patterns_descriptive_labels}
\end{figure}

The result is presented in Figure~\ref{fig:patterns_descriptive_labels}: The orange line is the growth curve for the types of each pattern. For example, in 1905, Mann had come across 78 different \textit{-nis} types. The blue line indicates the growth curves for the known types, which is a fraction of the growth curve for the input types (the scale for the input is on the left hand side of a panel, the scale for the output is on the right hand side, where applicable). In 1905, Mann had used 38 of the 78 types he had read, or 49\%. New words that Mann has probably not encountered in his readings are marked by an asterisk.
For example, Mann formed one new \textit{-tum} type in 1897, \textit{Bajazzotum}. 

The three unproductive patterns \textit{-nis}, \textit{-sam} and \textit{-sal} are clearly distinguishable from the other patterns: Their vocabulary growth curves have reached a limit, and additional input tokens should not change the final number much. The other patterns exhibit a linear growth. The patterns are still expanding, and texts that Mann read will regularly have contained new words of these patterns, never seen before. This is one of the hallmarks of productive patterns.

The degree of re-use of known words differs considerably among the patterns. On the one end, there is \textit{-nis}, with a recycle rate of 52\% (41 out of 79 types in the input are reproduced). On the opposing end, there is \textit{-tum}, with a 7\% rate (23 out of 335 types in the input are reproduced). The other patterns occupy middle ground, with a tendency of more productive patterns towards lower recycle rates.

In general, many hypotheses about the productivity of \textit{-tum}, \textit{-bar}, \textit{-los}, and \textit{-lich} and the unproductivity of \textit{-nis}, \textit{-sam}, and \textit{-sal} are conceivable: Apart from the recycle rate, it could be the number of cumulative types in the input, or the rate of hapaxes among the tokens (productivity in the narrow sense), or the rate of new types in the input among all input types \citep[P\textsubscript{neo} in][]{berg2020changes}. These and other hypotheses will be tested statistically in the next section.

\subsection{Statistical Modeling}

\noindent
We model the number of new words in Mann's own writings per year, with the following potential predictors for each pattern:
\begin{itemize}
    \item The number of types in the input in that year, log-transformed;
    \item the cumulative number of types in the input, log-transformed;
    \item the number of tokens in the input in that year, log-transformed;
    \item the number of hapaxes in the input in that year;
    \item the number of new types in the input in that year, log-transformed;
    \item the number of known types that Mann uses in his writings, log-transformed;
    \item the average distance between the embeddings of all types in the input that year and the centroid of all embeddings of all types encountered in the input so far;
    \item the ratio of hapaxes that year and the sum of all tokens that year (`productivity in the narrow sense');
    \item the ratio of new words that year and all types that year (`P\textsubscript{neo}');
    \item the ratio between the cumulative re-used types and the cumulative set of all types (`recycle rate');
    \item the year of reading/writing.
\end{itemize}

Each year in which Mann published a text represents one case for each pattern; this leads to a table of 18 years $\times $7 patterns, each with one response variable and 10 predictors. We analyzed these data using Generalized Additive Models (GAMs, Wood, 2017), which are able to capture non-linear relationships between predictors and response variables. 

We started with a full model with smooth terms for all variables plus an interaction term between centroid distance and the number of different known types Mann uses, and subsequently reduced the number of predictors if they were not significant at least at the 0.05 level. We use the log-transformed number of types that Mann uses as an offset in the model; after all, we are not interested in the absolute number of Mann's new formations, but as a fraction of the number of types of that pattern --- in years with long texts and many relevant types, we expect to see more new types compared to years with a lower output. We also tested whether predictors interact with each other; our criterion for model selection was based on Akaike's information criterion (AIC). The model with the lowest AIC has one smooth term (P\textsubscript{neo}) and one tensor product smooth (the interaction between the average distance of a pattern's embeddings from the pattern's centroid on the one hand, and the log-transformed number of reproduced types in Mann's output on the otehr hand). Both the smooth term and the interaction significantly predict the number of Mann's new formations (see Table~\ref{table:gamm_mann}).

\begin{table}[H]
    \centering
    \caption{GAMM fitted to the Mann data.  CentroidDistance = Average distance between a word's embedding and the centroid of the pattern. logKnownTypOut = log-transformed count of different lexemes in Mann's writings that were part of his readings up until that point.}
        \resizebox{\textwidth}{!}{
            \begin{tabular*}{\textwidth}{@{\extracolsep\fill}lrrrrrr}
            \hline
            {\bf Parametric coefficients} \\
            \hline
            Variable & Estimate  & Std. Error & {\it t}  &  {\it p} \\
            \hline
            Intercept & -4.3696   & 0.3126  & -13.98 & \texttt{<}.0001\\
            \hline
            {\bf Smooth terms}\\
            \hline
            Variable & edf  & Ref.df & F  &  {\it p}\\
            \hline
            P\textsubscript{neo}  & 0.931  & 9  & 13.57 & \texttt{<}.0001\\
            CentroidDistance*logKnownTypOut  & 3.104   & 24   & 10.67 & 0.0055\\
            \hline
            $R^2$ = .741\\
            \hline
            \end{tabular*}
}
\label{table:gamm_mann}
\end{table}

\begin{figure}
\begin{subfigure}[b]{\textwidth}
    \includegraphics[width=0.475\linewidth]{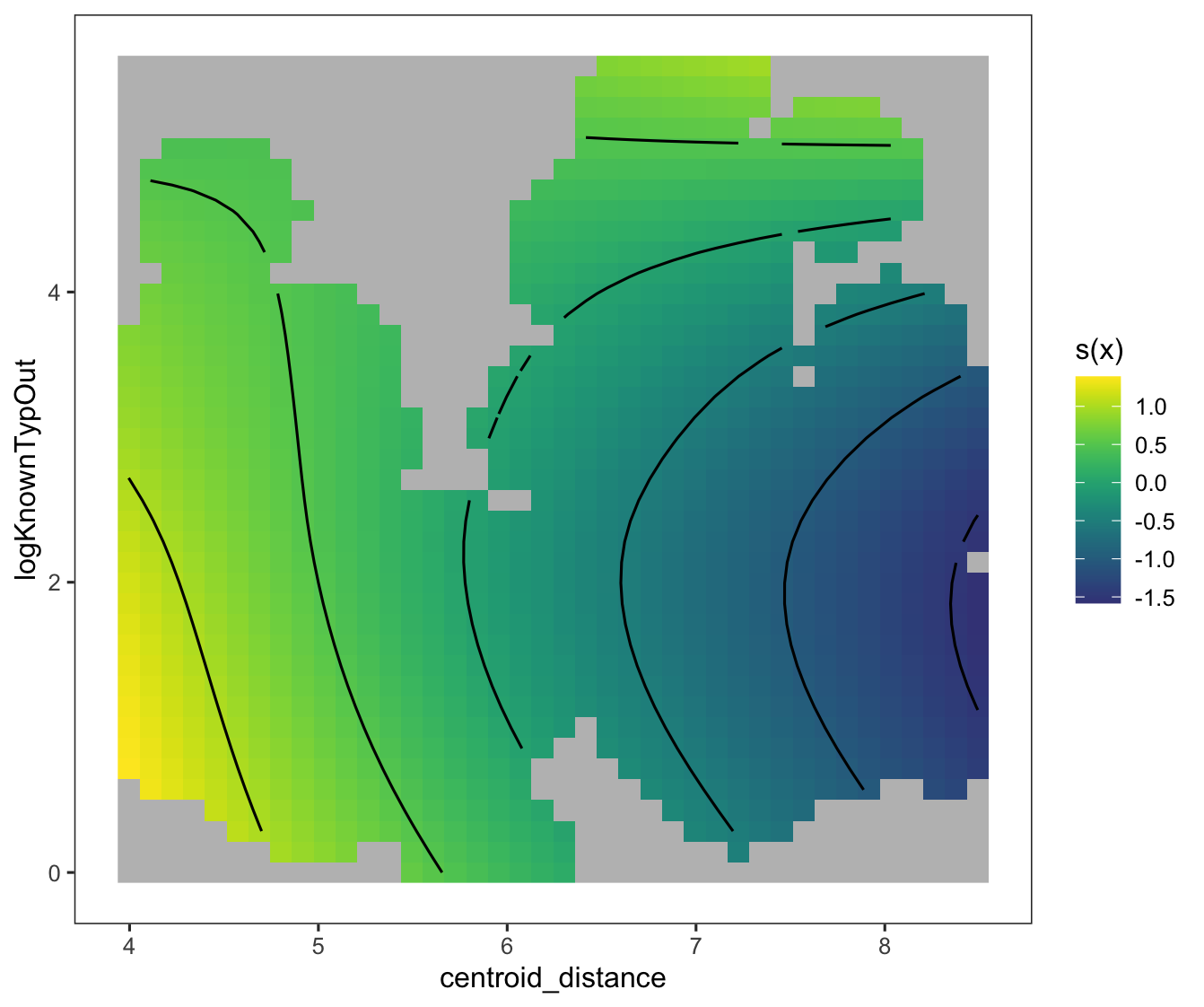}
    \hfill
    \includegraphics[width=0.475\linewidth]{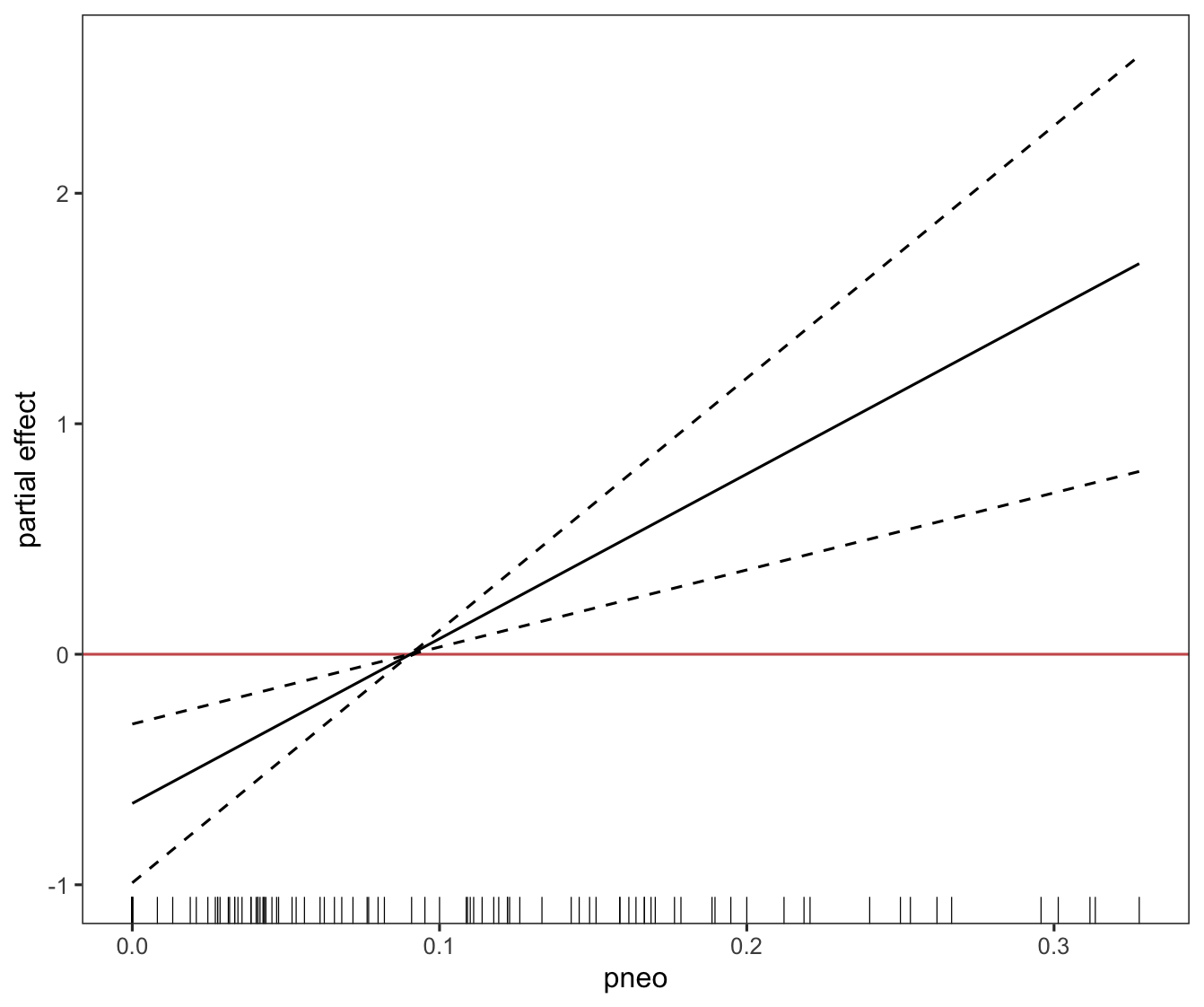}
\end{subfigure}
    \caption{Visualization of the model terms in the best GAM model discussed above: average centroid distance and the number of known types interact (left plot). Smooth of P\textsubscript{neo} as partial effect (right plot)}
    \label{fig:mann_gamm}
\end{figure}

The left panel of Figure~\ref{fig:mann_gamm} shows the interaction between the average distance from the centroid (horizontal axis) and the number of known types that Mann produces. Yellow tones indicate a higher probability for a new word. Decreasing semantic homogeneity (i.e., an increasing distance from the centroid) inhibits productivity. This seems plausible: The less homogeneous a pattern is, the harder it is to generalize it to new formations. This effect can (at least partly) be remedied by an increase in the number of known types in the output (top right part of the contour plot). In the right panel of Figure~\ref{fig:mann_gamm}, we observe an effect of P\textsubscript{neo}: The probability for a new formation increases with the ratio of new types among all types.  

\captionsetup[subfigure]{labelformat=empty}

In other words, the more known words Mann re-uses, and the more semantically homogeneous a pattern is, the more likely he is to go beyond what he has read. The same holds for the ratio of new words among all words: the more new words there are for any pattern relative to known words, the more likely Mann is to produce a new word himself.

From a cognitive perspective, this makes sense. The more types of a pattern somebody reuses, and the more similar they are, the easier it is to go beyond what has been read and heard so far. So easy, in fact, that Mann arguably did not perceive these formations as new, bold and/or odd. This leads to an interesting question: Given the words Mann has read, and the order he has read them in, and also given the meanings we extracted from the respective texts (using word2vec) --- can we construct a model of a mental lexicon to assess the `ease' of Mann's word formations? Of course not; but we can approximate it, and we set out to do so using the discriminative lexicon model introduced above in the remainder of this section.

\subsection{Modeling: DLM}

We trained various DLM models on pairings of lexemes and word embeddings, preserving the frequency with which the words appear in Mann's readings. Lexemes are represented as 4-grams, while the 300-dimensional Word2Vec vectors introduced above serve as word embeddings. All calculations were performed using the Julia package JudiLing \citep[][]{Heitmeier:Chuang:Baayen:2024}\footnote{The \textbf{JudiLing} package is in the julia repository, the latest development version can be downloaded from \url{https://github.com/quantling/JudiLing}.}, using both endstate learning and frequency-informed learning \citep{Heitmeier:Chuang:Axen:Baayen:2023}. 

The Word2Vec algorithm contains a frequency threshold (``min\_count''  in the gensim implementation) which was set to 5. A lower threshold leads to more unreliable embeddings, but is more inclusive of low-frequency words; a higher threshold leads to more robust embeddings, but contains fewer rare words. This is a dilemma for anyone who wishes to investigate morphological productivity (often manifest in rare words) with the means of context-independent word embeddings that are not given access to substrings within words; it means that the embeddings are the bottleneck for our model construction, limiting the number of different lexemes to the 42,966 different lexemes mentioned above. These lexemes occur 13 million times in Mann's readings between 1888 and 1915.

We trained two models every five years, one using endstate-of-learning (EOL) and one taking frequency of occurrence into account (frequency-informed learning, FIL). Each model was trained with all lexemes that had been read up until and including the year in question: The 1893 model, for example, contains all lexemes in Mann's reading up until 1893 (for which embeddings exist). Table~\ref{tab:mann_accuracies} reports the accuracies of the resulting models, both as a conservative measure (@1, the fraction of all form-meaning correspondences correctly reproduced by the model) and more lenient measure (@10, the fraction of all form-meaning correspondences where  the model's estimate is among the top 10 items). It is obvious that the models get worse over time: The more words there are, the harder it gets for a linear model to distinguish them; after all, between 1898 and 1903, the number of different lexemes more than doubles. It is also obvious that frequency-informed learning fares much worse than endstate-of-learning, when evaluated on accuracy @1 or accuracy @10 on a type basis.  

The endstate-of-learning models show what is theoretically possible with the data, given unlimited training time on the input dataset. The frequency-informed models, on the other hand, are more realistic for the situation at hand: They will learn frequent words well, and inevitably struggle with low-frequency words. 

Although at first sight the low type accuracies of FIL are surprisingly disappointing, these low accuracies are the inevitable consequence of taking a usage-based approach.   To see this, consider the consequence of the fact that in word frequency distributions, the hapax legomena constitute roughly 50\% of the types. Words that the model encounters only once cannot be learned given the barrage of tokens of high frequency words that the model is continuously confronted with.   It therefore makes sense to evaluate FIL-based models on what proportion of the tokens are properly recognized. Table~\ref{tab:mann_accuracies}  therefore lists accuracy @1 using token-based evaluation. These accuracies are substantially higher and, across all 5 measurement years, outperform type-based accuracy using endstate learning. In other words, the words that are useful to know because they occur frequently are learned much better by FIL-trained models.  

A further consideration to keep in mind is that it is unrealistic to expect a model of isolated word comprehension to be able to predict words perfectly: words typically occur in utterances, and the utterance will often be an excellent guide to a more precise understanding that cannot be achieved on the basis of the bottom-up input of just the word on its own. A FIL model is just a small part of a much more comprehensive set of language skills. 

Finally, the performance of FIL raises interesting questions about how well one can generalize from the high-frequency words belonging to a word formation pattern to the low-frequency words in that pattern.  Higher frequency words, and especially higher-frequency derived words, tend to have more senses and tend to have more semantic idiosyncracies.  Thus, the more a word formation pattern is dominated by high-frequency words, the less one may expect generalization to low-frequency words in that pattern to be successful.  The category-conditioned productivity measure (the Good-Turing ratio of hapax legomena to tokens) captures this by penalizing, however crudely, for token frequencies.

\begin{table}[ht]
    \centering
    \caption{Comprehension accuracies for training data. Accuracies @1 pertain to perfect recognition, accuracies @10 to accuracy evaluated as belonging to the top 10 nearest neighbors of the intended meaning. Tokens and types refer to token-based and type-based evaluation.}
    \label{tab:mann_accuracies}
\begin{tabular}{lrrrr} \hline 
  learning &  @1 types  & @1 tokens   & @10 types  & \#lexemes \\ \hline
  1893, EOL     &  0.809 & - & 0.929   & 15,219\\
  1893, FIL     &  0.121 & 0.819 & 0.238  & \\ 
  \hline  
  1898, EOL     &  0.649 & -  & 0.815  & 33,777 \\
  1898, FIL     &  0.062 & 0.826 & 0.125  &  \\ 
  \hline  
  1903, EOL     &  0.625 & -  & 0.792  & 38,606 \\
  1903, FIL     &  0.056 & 0.822 & 0.112  & \\ 
  \hline  
  1908, EOL     &  0.615 & - & 0.78  & 41,711 \\
  1908, FIL     &  0.052 & 0.817 & 0.104  &  \\ 
  \hline  
  1913, EOL     &  0.614 & -  & 0.777 & 42,747 \\
  1913, FIL     &  0.051 & 0.815 & 0.102 &   \\ 
  \hline  
  
\end{tabular}
\vspace{0.5em} 
\end{table}

In what follows, we make use of FIL-based models, firstly because they are more in line with usage-based linguistics, and second, because we expect to find structure in the model's association values, even though model predictions for individual lower-frequency types will lack precision. Specifically, in the following, we use the first FIL model in Table~\ref{tab:mann_accuracies} (1893) and ask, for each pattern of interest, how well these patterns have been learned: What is the accuracy for all \textit{-nis} words in the training set, for example?

Table~\ref{tab:mann_accuracies_training} shows that the accuracy varies greatly between the patterns. The level of accuracy for individual patterns is higher than the accuracy of the whole model, save for \textit{-los}. Infrequent patterns (such as \textit{-sal}) and frequent patterns (such as \textit{-lich}) have surprisingly high accuracies which are unexpected given their characterizations as unproductive or less productive. What we are seeing is an effect of the frequency distribution within each pattern. Generally, the more tokens per type there are, the better the model learns the pattern's words (Spearman's $\rho $ between acc@1 and tokens per type: 0.829, p = 0.021). That means the FIL model favors patterns that are high in tokens but low in types --- in other words, unproductive patterns, exactly as expected. Accordingly, token-based accuracy (@1 tokens) is highest for the unproductive patterns. This then would suggest that comprehension accuracy for training data in an FIL model cannot be predictive for a pattern's productivity, conceptualized as the capacity to generalize of a word formation pattern to unseen types.

\begin{table}[ht]
    \centering
    \caption{Comprehension accuracies for training data, model 1893, using FIL, for the individual word formation patterns.}
    \label{tab:mann_accuracies_training}
\begin{tabular}{lrrrrrr} \hline 
  learning &  @1 types    & @10 types & @1 tokens   &\# types & \# tokens & tokens per type\\ \hline
  -sam     & 0.272 & 0.546 & 0.909 & 22 & 286 & 13\\
  \hline  
  -nis     &  0.27  & 0.324 & 0.806 & 37 & 397 & 10.7\\
  \hline 
  -sal     & 0.25 & 0.25  & 0.903  & 4 & 72 & 18\\
  \hline  
  -bar     &  0.207  & 0.31 & 0.769  & 29 & 242 & 8.3\\
  \hline  
  -lich     &  0.158  & 0.288 & 0.752 & 287 & 3,047 & 10.6\\
  \hline  
  -tum    &  0.158  & 0.316 & 0.782  & 19 & 142 & 7.5\\
  \hline   
  -los     &  0.058  & 0.161 & 0.415 & 87 & 234 & 2.6\\
  \hline  
\end{tabular}
\vspace{0.5em} 
\end{table}

This conclusion is supported by an examination of how well the FIL model predicts held-out data.
Usually, models are validated by carefully splitting the data into training and held-out data (as we did above in section~2). In the case of the Mann corpus, however, we know exactly which new words he encountered after 1893, and we will use these words (grouped into their respective patterns) as validation data. For example, there are 35 \textit{-lich} words that Mann encountered after 1893, among them formations like \textit{erzbischöflich} ('archbishop-ly') or \textit{weltanschaulich} ('ideologically'). For each of these words, we let the model predict an embedding, and we then compare this predicted embedding with the empirical Word2Vec embedding. Table~\ref{tab:mann_accuracies_held_out} shows that the FIL-based model trained on data until 1893 does not generalize beyond the training data. For none of the held-out words, the empirical embedding comes in among the ten most similar to the predicted one. Even if we keep in mind that  each predicted embedding is evaluated against the full set of 15,210 emprical embeddings, prediction accuracy for FIL is underwhelming.  

However, we can lower the accuracy threshold to the set of the 100 or 1000 most similar embeddings, and then some structure becomes visible. The patterns that have the lowest number of tokens per type in Table~\ref{tab:mann_accuracies_training} have the highest accuracy values; both are in fact significantly negatively correlated (Spearman's $\rho -0.943, p = 0.005$). So while the FIL model has difficulties learning words from productive patterns, it is able to learn something about the pattern itself, which in turn enables it to understand new words slightly better than words from more unproductive patterns.  But precise prediction for held-out data is clearly out of reach.

\begin{table}[ht]
    \centering
    \caption{Comprehension accuracies for held-out data (Types), model 1893, FIL.}
    \label{tab:mann_accuracies_held_out}
\begin{tabular}{lrrrr} \hline 
  learning &  @10     & @100 & @1000  & \#new words >1893 \\ \hline
  -los     &  0.0  & 0.024 & 0.342 & 41 \\
  \hline  
  -lich     &  0.0  & 0.143 & 0.257  & 35\\
  \hline  
  -bar     &  0.0  & 0.0  & 0.273 &11 \\
  \hline  
  -tum    &  0.0  & 0.1  & 0.3 & 10 \\
  \hline  
  -sam     & 0.0 & 0.0 & 0.25 & 4 \\
  \hline  
  -nis     &  0.0  & 0.0 & 0.0 & 3 \\
  \hline  
\end{tabular}
\vspace{0.5em} 
\end{table}

Above, we pointed out that training a model with FIL implies giving high-frequency words the opportunity to dominate learning.  We anticipated adverse effects for generalization to held-out data.  This expectation was borne out, but there is further snag that is worth discussing. 

Recall that 
the Word2Vec algorithm needs a certain minimum amount of occurrences to generalize over a word's co-occurrences and construct high-quality embeddings. In our case, we set the minimum value to 5. This in turn means that we exclude nonce formations, the hallmark of productive patterns. These nonce formations are in general more transparent (in the sense of compositional) than frequent formations, which tend to be charged with any kind of additional semantic, syntactic, or pragmatic information. We expect the DLM model to predict infrequent --- and semantically regular ---- formations particularly well, but sadly, we have no embeddings for these words. If we had, the accuracy values for the FIL model (Table~\ref{tab:mann_accuracies_held_out}) would arguably increase markedly (after all, we are missing out on 255 \textit{-lich} types that occur less than 5 times in the data set, such as  \textit{österlich} 'Easterly').  

In summary, not only is the model trained without low-frequency words, it is also tested on words that are not the lowest frequency words. Given the Zipfian nature of word frequency distributions, most of the `low-frequency' words that occur more often than 5 times likely occur much more often than 5 times. 

Due to the currently unavoidable limitations of our data, the FIL model is confronted with the task of generalizing from a dataset with systematicities that are much more limited than we had originally hoped.  How severe these limitations are becomes apparent when we replace learning with FIL with learning with a deep neural network with a hidden layer of 1000 RELU units, with early stopping to avoid overfitting. 
This model achieves 99.5\% accuracy on the training data, but is not much better at predicting the held-out data. Accuracy @10, for example, is zero for all patterns except for \textit{bar}, which clocks at 9\%).  As can be seen in Table~\ref{tab:mann_deep_learning}, the deep model gets most of the handful of novel types for \textit{-sam} and \textit{-nis} correct, but for the other word formation patterns, accuracy @1000 is only slightly higher than for FIL.  As this deep learning network is not frequency aware, and as a more powerful version of endstate learning informs us about what is learnable in the limit of experience, we can conclude that the present dataset is too unsystematic to afford precise prediction.

\begin{table}[ht]
    \centering
    \caption{Comprehension accuracies for held-out data (types), model 1893, deep mapping.}
    \label{tab:mann_deep_learning}
\begin{tabular}{lrrrr} \hline 
  learning &  @10     & @100 & @1000  & \# new words $>$ 1893 \\ \hline
  -los     &  0.0  & 0.073 & 0.463 & 41 \\
  \hline  
  -lich     &  0.0  & 0.057 & 0.371 & 35 \\
  \hline  
  -bar     &  0.09  & 0.09  & 0.363 & 11 \\
  \hline  
  -tum    &  0.0  & 0.2  & 0.3 & 10 \\
  \hline  
  -sam     & 0.0 & 0.0 & 1 & 4 \\
  \hline  
  -nis     &  0.0  & 0.0 & 0.667 & 3\\
  \hline  
\end{tabular}
\end{table}

One possible way to address this problem would be to measure the model's accuracy not against the empirical embeddings, but against a constructed embedding obtained by adding the centroid embedding to the embedding of the base word, see, e.g., \citet{Baayen:Chuang:Shafei:Blevins:2019} and \citet{Heitmeier:Chuang:Baayen:2024}. All words that are currently excluded because their frequency is too low to obtain a reliable embedding can then be included with a constructed embedding. These constructed embeddings are likely too regular, and therefore err on the side of semantic transparency. Nevertheless, their inclusion would allow us to test to what extent inclusion of the low-frequency words will equip FIL with improved prediction accuracy for held-out words.  Given the overwhelming effect of high-frequency words in FIL learning, it is far from self-evident that prediction accuracy will indeed improve. If the answer is negative, this suggests that research on productivity in a usage-based perspective will have to focus far more on how usage in utterances supports the interpretation of novel words, using contextualized embeddings. We leave this question for further research.

One final question we would like to address in this section is, what is the FIL model learning actually? Even if the accuracy values are low, some of the pattern's aspects are learned, as witnessed by the close correlation between tokens per type and accuracy. Where is the information about, for example, \textit{-nis} words stored in a model that makes do without words, patterns, rules and constraints? The answer, as in section~2 above, is: at least partly in the pattern's 4-grams.

Table~\ref{tab:top5_ngram_mann_EOL} shows the top five strongest correlations between the row vectors in $\bm{F}$ and the centroid embeddings of the seven derivational patterns discussed in this section. Each centroid, save for the most type-frequent pattern \textit{-lich}, is most strongly associated with its respective 4-gram in word-final position.

\begin{table}[H]
\centering
\caption{Top five strongest correlations between the row vectors in $\bm{F}$ and the centroid embeddings of the seven derivational patterns in an FIL model of Mann's readings until 1893. Coefficients highlighted in red represent the strongest correlation between the row vectors of the 4-gram and the embeddings of the centroid.
}
\label{tab:top5_ngram_mann_EOL}
{\footnotesize
\begin{tabular}{r|rrrrrrrrr} \hline
 & \textsc{-bar} & \textsc{-lich} & \textsc{-los} & \textsc{-nis} & \textsc{-sal} & \textsc{-sam} & \textsc{-tum} \\ \hline
  bar\# & {\color{red}\textbf{0.82}} & 0.704 & - & - & - & 0.572 & -\\ 
  tig\# & 0.671 & {\color{red}\textbf{0.772}} & - & - & - & 0.575 & -\\ 
  dig\# & 0.628 & 0.709 & - & - & - & - & -\\ 
  ial\# & 0.57 & .0634 & - & - & - & - & -\\ 
  siv\# & 0.547 & 0.605 & - & - & - & - & -\\ 
  los\# & - & - & {\color{red}\textbf{0.685}} & - & - & - & -\\ 
  nlos & - & - & 0.642 & - & - & - & -\\ 
  tlos & - & - & 0.62 & - & - & - & -\\ 
  \#unb & - & - & 0.602 & - & - & - & -\\ 
  utal & - & - & 0.6 & - & - & - & -\\ 
  nis\# & - & - & - & {\color{red}\textbf{0.949}} & - & - & -\\
  ung\# & -  & - & - & 0.841 & - & - & -\\
  gabe & -  & - & - & 0.713 & - & - & -\\
  ion\# & -  & - & - & 0.703 & - & - & 0.78\\
  tnis & -  & - & - & 0.703 & - & - & -\\
  sal\# & -  & - & - & - & {\color{red}\textbf{0.894}} & - & -\\
  ksal & -  & - & - & - & 0.878 & - & -\\
  icks & -  & - & - & - & 0.878 & - & -\\
  cksa & -  & - & - & - & 0.876 & - & -\\
  hick & -  & - & - & - & 0.751 & - & -\\
  sam\# & -  & - & - & - & - & {\color{red}\textbf{0.891}} & -\\
  tsam & -  & - & - & - & - & 0.554 & -\\
  egt\# & -  & - & - & - & - & 0.536 & -\\
  tum\# & -  & - & - & - & - & - & {\color{red}\textbf{0.859}}\\
  mus & -  & - & - & - & - & - & 0.742\\
  tät\# & -  & - & - & - & - & - & 0.742\\
  ität & -  & - & - & - & - & - & 0.715\\
  \hline
  \label{table:top5_ngram_mann_FIL}
\end{tabular}
}
\end{table}

This demonstrates that the model is on the right track: It has learned that the part of the word that carries the pattern's letter chunk is connected to the pattern's semantic contribution. Closer inspection reveals a more intricate knowledge of German derivational morphology:

\begin{itemize}
    \item Among the top correlations of adjectival \textsc{-bar} and \textsc{-lich} are 4-grams containing other adjectival patterns (native \textit{-ig} and foreign \textit{-al} and \textit{-iv}).
    \item For the privative adjectival pattern \textsc{-los}, one of the most strongly correlated forms (apart from \textit{-los} in three variations) is the negative prefix \textit{un-}.
    \item \textsc{-nis} forms action nouns or nouns that denote the result of a dynamic action. Among the strongest correlations for the centroid are two other nominal patterns with a similar function, native \textit{-ung} and foreign \textit{-ion}.
    \item Nominal \textsc{-tum} is most similar to foreign nominal patterns \textit{-ität}, \textit{-ion}, and \textit{-ismus}, all of which are partially overlapping semantically.
    \item \textsc{-sal} is particularly interesting because the highest correlations involve bits of \textit{Schicksal} 'fate', by far the most frequent of the few \textit{-sal} words. The pattern is not related to other patterns, possibly because it is not recognized as a pattern by the model.
\end{itemize}

In all these cases, it is the semantic and/or syntactic overlap between the centroids of different patterns which leads to the observed correlations. It is worth stressing, though, that these correlations emerge without explicitly providing any morphological and/or lexical information.  Considered jointly, we conclude that the FIL model, in spite of the limitations of our dataset, has learned a lot about the relations between form and meaning for the German word formation patterns in our Mann corpus.

\section{General discussion}

In this study, we have used the Discriminative Lexicon Model (DLM) as a tool for probing morphological productivity.  Our main findings are the following. First, starting out from the fact that without systematicity, machine learning and human learning  cannot generalize to previously unseen, `held-out', data, we formalized the degree of productivity of a word formation pattern as the prediction accuracy of the DLM for held-out data.   For Finnish nominal inflection for case and number, the predictions of the DLM align well with the different degrees of productivity of the many inflectional classes of Finnish, as gauged with classical well-established measures.  For prefixal derivation in Malay, accuracies for held out data decreased, unsurprisingly as the semantics of derived words typically are less transparent than the semantics of inflected words.  For English compounds, accuracy for held out data was the lowest, which is also unsurprising as the meanings of novel compounds can be notoriously difficult to predict \citep[see, e.g.,][]{schafer2020constituent}, and given that there are no systematic form-meaning parallels (which led \citet{Schultink:62} to argue that compounds do not form a morphological category).  Since in terms of type counts, compounding is extremely productive, we conclude that systemic productivity can be assessed well with the DLM, but that the productivity of compounding is outside the scope of the model, as this type of word formation is made possible by onomasiological tinkering (`bricolage').  

Second, we have shown by means of an examination of what Thomas Mann is likely to have read, and what he wrote, that the rate at which Mann produces novel derived words is extremely low.   There are far more novel words in his input than in his output.  We document a strong correlation between ($\rho = 0.95$) the rate of novel words in Mann's input and the rate of novel words in his output.  Mann's comprehension productivity (assuming that he understood what he read) was clearly much higher than his production productivity.    This asymmetry is also well documented in the acquisition literature \citep[see, e.g.,][ and references cited there]{gershkoff2013word} and emerges time and again in the DLM \citep{Chuang:Bell:Banke:Baayen:2020,Heitmeier:Chuang:Baayen:2024}.  
However, the magnitude of the difference in comprehension and production productivity for a prolific and gifted writer such as Thomas Mann surprised us. But it fits well with existing studies investigating the spread of novel words \citep{de2012ash} suggesting that innovations spread through social networks, starting with a single innovator, being adopted by followers, and with bursts of popularity possibly standing in the way of entrenchment in a community's lexicon \citep{chesley2010predicting}.

Third, across all case studies, the centroid in semantic space of all words sharing a morphological pattern turned out to play an important role.  When a DLM model is set up with linear mappings between form and meaning, the row vectors of the comprehension mapping and the column vectors of the production mapping specify semantic vectors.  It turns out that the more an n-gram overlaps with an exponent, the stronger its vector in the mappings correlates with the centroid of the embeddings of the words sharing that exponent.  The centroids are the `average' meanings of the exponents, and the linear mappings converge on these centroids as the best estimates for the semantics of sublexical n-grams.  We document this across all our datasets: Finnish nominal inflection, Malay prefixation, and the derivational suffixes in the writings of Thomas Mann. For Malay, the correlation of the embedding of a derived word with the centroid of its morphological pattern emerges as both a measure of morphological transparency and a robust predictor of lexical decision times.  And Thomas Mann is less likely to produce a novel derived word with a given suffix the greater the average distance is of the embeddings of all derived words to the corresponding centroid. In other words, the less transparent the word formation pattern, the less its production productivity.
One of the advantages of computational modeling is that the consequences of theoretical decisions are clearly visible. In the present study, the simulation experiments with endstate learning (EOL) and deep learning provide insight into what can be learned with infinite cycles through the dataset of word types.  Frequency-informed learning (FIL) implements a usage-based alternative. Training mappings with FIL shows that words that occur once only cannot be learned --- a result that will not surprise anyone with experience in machine learning. Since in word frequency distributions, the hapax legomena often make up around 50\% of the types, type accuracies for FIL are unavoidably low.  On the other hand, since high-frequency words are learned well, the proportion of tokens that is learned correctly is much higher.    Nevertheless, a FIL model for Mann trained on the materials from before 1893 struggles with understanding the novel forms encountered after 1893, although it may get the gist correct. Deep learning does not fare much better. The reason for this is the limitations of our dataset that come with the word2vec embeddings that we used. These embeddings were calculated from Mann's reconstructed own experience using word2vec. In order to obtain reasonably reliable embeddings, embeddings were calculated only for words with a frequency equal to or greater than 5.  As a consequence, both the training data and the test data are biased towards more frequent, and hence less semantically predictable, words. Unsurprisingly, precise prediction suffers dearly.  Nevertheless, it is clear from an inspection of how pattern centroids are represented in the comprehension mapping that, first, the FIL model has already learned a lot about the general semantics but also that, second, the centroids of semantically similar affixes, as estimated from our dataset, are similar and are confusable. An open question is whether including constructed `regularized' embeddings for the lowest-frequency words in the Mann corpus will enable FIL to generate more precise predictions for held-out data.

This kind of problem is not specific to our case study of Thomas Mann. The studies of Finnish inflection and Malay derivation make use of corpus frequencies, which are estimates of community usage. As a consequence, for individual language users, the experience with the lowest frequency words is severely overestimated, and experience with higher-frequency words somewhat underestimated.  As a consequence, the models are unavoidably biased.

An advantage of computational modeling with the DLM is that this theoretical framework does not require the analyst to make discrete decisions about whether words are exceptional or regular, as in the theory of \citet{Yang:2016}. According to his tolerance principle, a rule is completely unproductive when the number of types $V$ divided by the natural logarithm of $V$ is greater than the number of types that are exceptional in some way. Otherwise, it is completely productive.  But what counts as an exception?  In Finnish, the semantics of plurality vary systematically across case \citep{nikolaev:2022}, and within case-number combinations, individual words again show considerable variation.   Does this semantic diversification make the whole inflectional system unproductive? Furthermore, at the form side, how to properly set up the inflectional classes of Finnish nouns is far from clear, the current standard being a compromise between lumping and splitting \citep{Nikolaev:Chuang:Baayen:2025}.  The DLM does not force the analyst to make any such ad hoc decisions. The model is not informed about the inflectional classes, nor about case and number. It is provided with rich (but not perfect) information about words' meanings, and it sets up form embeddings using simple n-grams.  The mappings between form and meaning optimize for prediction, but at the same time learn to remember the idiosyncracies of individual words' forms and meanings.  This approach offers novel ways of conceptualizing productivity (generalization as opposed to bricolage), and of assessing degrees of productivity for word formation patterns and associated allomorphs, for inflection and inflectional classes, and for compounding.  As we have shown, this approach also helps clarify the challenges for the analyst wishing to understand productivity at the level of the individual.    
Our ideal for productivity research is a much more ambitious simulation model in which virtual agents with DLM-like  mental lexicons are interacting over time in socially stratified communities. Such a model would make it possible to simulate and hopefully predict how innovations might spread in communities, and how productivity changes over the lifetime of both individuals and speaker communities. While such a full model of speakers within their groups within a language community is currently out of reach, studies that focus on simplified aspects are conceivable and may yield interesting and novel results. For example, if we equip a small number of agents with a toy DLM lexicon and  have them randomly produce „texts“ based on their mappings, and every agent „reads“ every other agent’s texts, we may observe changes to the individual mappings over time that result in change on the level of the community, as measured by output texts. If these changes turn out to be similiar to actual language change, this would both explain the mechanics of language change in a new way, and indicate that an agent-based DLM approach is on the right track.

\newpage

\noindent
We are indebted to Alexandre Nikolaev for his feedback on this study.

\newpage

\noindent
\textbf{Appendix} \\ \ \\
\appendix

\section{Specifics of the CAOSS model}

\begin{figure}[hb]
    \centering
    \includegraphics[width=0.49\linewidth]{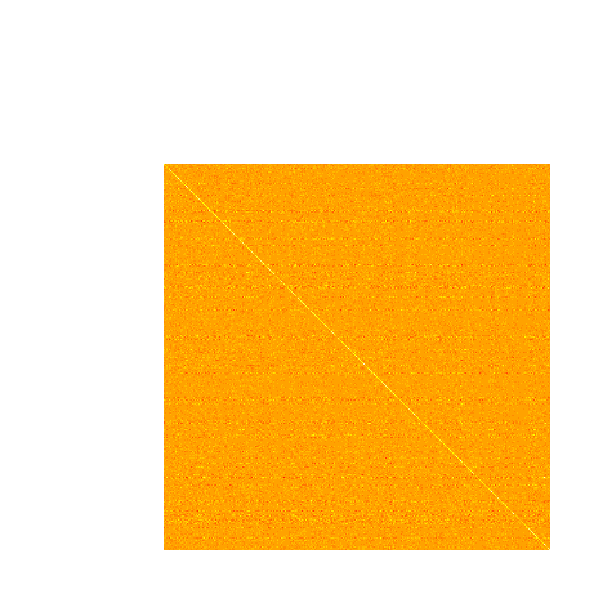}\hspace*{0.1em}
    \includegraphics[width=0.49\linewidth]{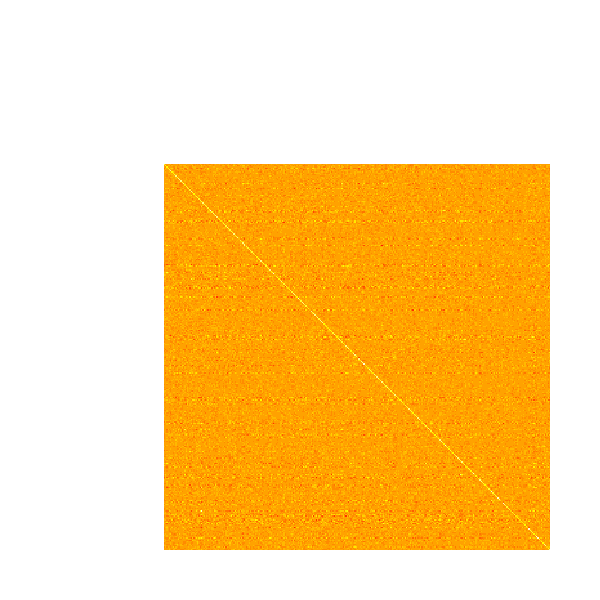}
    \caption{CAOSS mapping matrices, estimated for the dataset with 8147 English compounds.}
    \label{fig:CAOSS}
\end{figure}

Let $\bf{L}, \bm{R}$ and $\bm{C}$ denote the embeddings of the left constituents, the right constituents, and the compounds respectively. The
CAOSS model sets up a  linear mapping $\bm{M}$ by solving
$$
[\bf{L}\bf{R}] \bm{M} = \bm{C}.
$$
The mapping $\bm{M}$ consists of two blocks, one transforming $\bm{L}$, and one transforming $\bm{R}$  The output of these transformations are added to obtain $\bm{C}$:
$$
[\bm{L} \bm{R}] = \left[\begin{array}{c}
     \bm{M}_{L}  \\
     \bm{M}_{R} 
\end{array}\right] \bm{C}.
$$
Figure~\ref{fig:CAOSS} clarifies that the highest values of $\bm{M}_L$ and $\bm{M}_R$ are found on the diagonal. If all off-diagonal elements were exactly zero, these matrices would be identity matrices, and their joint effect would be simply addition of the constituent embeddings:
$$
\hat{\bm{C}} = \bm{L}+\bm{R}.
$$
Figure~\ref{fig:CAOSS} shows that CAOSS is finding a solution that is close to, but not identical to,  simple vector addition.

\newpage

\bibliographystyle{apalike}

\end{document}